\documentclass[11pt]{article}

\usepackage[final]{acl}

\usepackage{times}
\usepackage{latexsym}

\usepackage[T1]{fontenc}

\usepackage[utf8]{inputenc}

\usepackage{microtype}

\usepackage{inconsolata}

\usepackage{graphicx}

\usepackage{microtype}
\usepackage{hyperref}
\usepackage{url}
\usepackage{booktabs}
\usepackage{lineno}

\definecolor{darkblue}{rgb}{0, 0, 0.5}
\hypersetup{colorlinks=true, citecolor=darkblue, linkcolor=darkblue, urlcolor=darkblue}

\usepackage[utf8]{inputenc} 
\usepackage[T1]{fontenc}    
\usepackage{hyperref}       
\usepackage{url}            
\usepackage{booktabs}       
\usepackage{nicefrac}       
\usepackage{microtype}      
\usepackage{xcolor}         
\usepackage{amsfonts}       
\usepackage{amsmath}
\usepackage{caption}
\usepackage{subfigure}
\usepackage{subcaption}
\usepackage{multirow}
\usepackage{adjustbox}
\usepackage{colortbl}
\usepackage{array}
\usepackage{color}
\usepackage{graphicx}  
\usepackage{bbding}
\usepackage{amssymb}
\usepackage{pifont} 
\usepackage{wrapfig}
\usepackage{tcolorbox}
\usepackage{fontawesome5}
\usepackage[table]{xcolor}
\usepackage{adjustbox}
\usepackage[dvipsnames]{xcolor}
\usepackage{pifont}

\definecolor{blgrey}{rgb}{0.6,0.6,0.6}
\definecolor{bblue}{rgb}{0.855,0.933,0.98}
\definecolor{dblue}{HTML}{5297D6}
\definecolor{gainred}{rgb}{0.1,0.5,0.3}
\definecolor{citecolor}{HTML}{0071BC}
\definecolor{linkcolor}{HTML}{ED1C24}
\definecolor{dkgreen}{rgb}{0,0.6,0}
\definecolor{gray}{rgb}{0.5,0.5,0.5}
\definecolor{mauve}{rgb}{0.58,0,0.82}
\usepackage{listings}
\usepackage{caption}
\usepackage{wrapfig}
\usepackage{algorithm}
\usepackage{algpseudocode} 
\usepackage{float}
\usepackage{lipsum} 
\usepackage{tabularx}
\usepackage{enumitem}  %
\usepackage{makecell}
\usepackage{pifont} %
\usepackage{algpseudocode}
\usepackage{algorithm}
\usepackage{multirow}
\usepackage{xcolor}
\usepackage{booktabs}

\tcbuselibrary{breakable}

\newcommand{\cmark}{\ding{51}}                           
\newcommand{\xmark}{\textcolor{lightgray!150}{\ding{55}}}

\title{Think, Act, Build: An Agentic Framework with Vision Language Models for Zero-Shot 3D Visual Grounding}

\author{
    Haibo Wang$^\blacklozenge$, 
    Zihao Lin$^\blacklozenge$, 
    Zhiyang Xu$^\lozenge$, 
    Lifu Huang$^\blacklozenge$ \\
    $^\blacklozenge$University of California, Davis \quad $^\lozenge$Virginia Tech \\
    \texttt{\{hibwang, lfuhuang\}@ucdavis.edu}
}

\begin{document}
\maketitle
\begin{abstract}
3D Visual Grounding (3D-VG) localizes objects in 3D scenes via natural language descriptions. Current zero-shot Vision-Language Models (VLMs) typically rely on preprocessed 3D point clouds and pretrained 3D detectors, reducing grounding to static proposal matching. To address this, we propose "\textbf{\textit{Think, Act, Build (TAB)}}", a dynamic agentic framework that reformulates 3D-VG as a 2D-to-3D reconstruction paradigm operating directly on raw RGB-D streams. TAB decouples the task by leveraging 2D VLMs for complex spatial semantics and multi-view geometry for 3D structure instantiation. 
Guided by a specialized skill, our VLM agent dynamically invokes visual tools to track targets across 2D frames. To overcome multi-view coverage deficits inherent in strict VLM semantic tracking, we introduce Semantic-Anchored Geometric Expansion. This mechanism anchors the target in a reference video clip and uses multi-view geometry to propagate its spatial location across unobserved frames, mapping 2D visual cues to 3D coordinates.
Additionally, we manually refine existing benchmarks to eliminate reference ambiguity and category errors. Extensive experiments on ScanRefer and Nr3D demonstrate that TAB, relying entirely on open-source models, significantly outperforms previous zero-shot methods and even supervised baselines. Codes will be available at \href{https://github.com/WHB139426/TAB-Agent}{https://github.com/WHB139426/TAB-Agent}.
\end{abstract}

\section{Introduction}
3D Visual Grounding (3D-VG) \citep{achlioptas2020referit3d, chen2020scanrefer} is a fundamental task in 3D scene understanding, requiring an AI system to precisely localize a target object within a 3D physical space based on a free-form natural language query. This capability serves as a cornerstone for advanced applications such as human-robot interaction \citep{kim2024openvla}, embodied AI navigation \citep{anderson2018vln}, and AR/VR \citep{hoenig2015mixed}. Historically, the dominant paradigm in 3D-VG has relied on fully supervised learning frameworks \citep{roh2022languagerefer, zhao20213dvg, zhu2024scanreason, zhu2025llava3D}. While these methods achieve remarkable accuracy, their success is predicated on massive amounts of high-quality, densely annotated 3D vision-language datasets. The prohibitive cost and labor-intensive nature of collecting such 3D annotations inherently limit the scalability of supervised methods and their generalization to open-world, open-vocabulary scenarios.

\begin{figure*}
    \centering
    \includegraphics[width=1\linewidth]{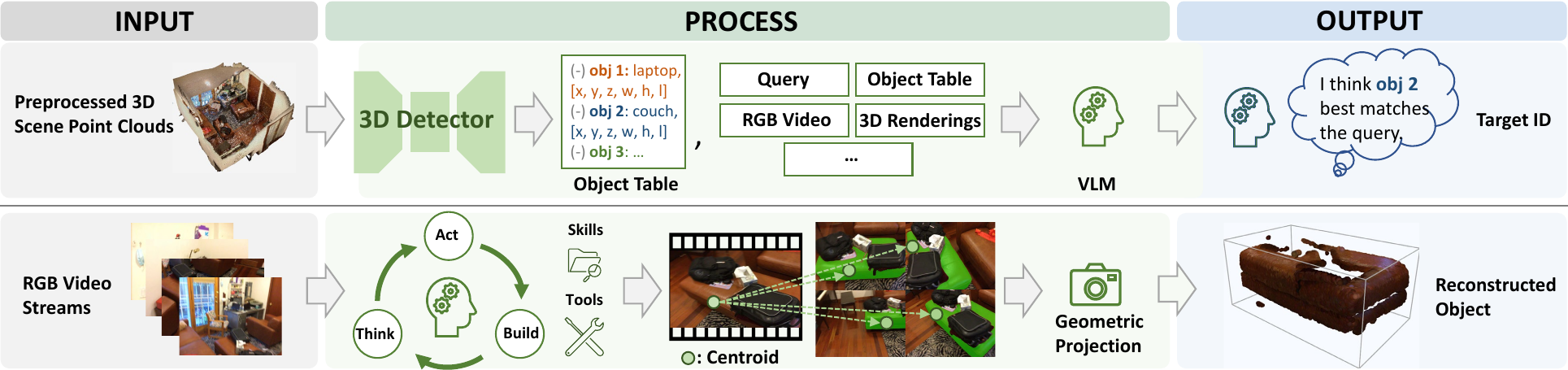}
    \caption{\textbf{(1) Top:} Previous methods rely on preprocessed 3D point clouds, degrading the task into proposal matching. \textbf{(2) Bottom:} Our \textbf{\textit{TAB}} operates directly on RGB-D streams. Through an iterative Think-Act-Build process, the agent reconstructs the target object.}
    \label{fig:intro}
    \vspace{-10pt}
\end{figure*}

To circumvent the scarcity of 3D annotations, recent efforts have shifted towards zero-shot 3D-VG paradigms \citep{Yang2023LLMGrounder, yuan2024zsvg3d, zhang2024agent3d0, xu2025vlmgrounder, mi2025languagetospace}. By harnessing pre-trained LLMs \citep{touvron2023llama, qwen3.5} and VLMs \citep{Qwen3-VL, li2024llavaonevision, deitke2025molmo}, these methods can ground objects without scene-specific 3D training. However, existing frameworks encounter limitations when applied to streaming RGB-D settings.
First, the majority of current zero-shot methods (e.g., SeeGround \citep{li2025seeground}, SPAZER \citep{jin2025spazer}, SeqVLM \citep{lin2025seqvlm}) heavily rely on well-preprocessed 3D point clouds as inputs. By utilizing static 3D maps to pre-extract candidate bounding boxes, they degrade 3D-VG into a mere “proposal matching” classification task that restricts the VLM to simply selecting from a pre-defined pool of 3D proposals, rendering them ineffective in environments where preprocessed 3D point clouds are unavailable.
Second, while some methods attempt to operate directly on 2D images (e.g., VLM-Grounder \citep{xu2025vlmgrounder}), they rely exclusively on heuristic 2D semantic matching to associate multi-view observations. By failing to exploit the deterministic 3D geometry inherent in continuous video streams, their tracking becomes highly brittle under extreme viewpoint variations, inevitably yielding fragmented and inaccurate 3D geometries.

To overcome these limitations, we propose \textbf{\textit{Think, Act, Build (TAB)}}, an agentic framework that reformulates zero-shot 3D-VG from a static proposal matching process into a semantic reasoning and geometric reconstruction process without relying on preprocessed point clouds or 3D detectors. Our core motivation stems from the insight that 2D VLMs excel at complex reasoning, leaving the precise 3D structural instantiation entirely to deterministic multi-view geometry. As shown in Figure \ref{fig:intro}, by directly operating on sequential RGB-D streams, \textbf{\textit{TAB}} bridges semantic intent with physical space by leveraging VLMs to ground the target into fine-grained 2D masks and then geometrically reconstructing these masks into a 3D object. Rather than adhering to a rigid pipeline, the framework orchestrates a dynamic, iterative process guided by an expert \textbf{\textit{3D-VG Skill}}. Following a ReAct-style paradigm  \citep{yao2022react}, the agent iteratively ``Thinks'' by reasoning over the \textbf{\textit{3D-VG Skill}} blueprint and current visual context to plan the next step, and ``Acts'' by invoking specialized tools (e.g., 2D detectors and segmenters) to interact with the visual environment to yield necessary observations for 3D-VG. Crucially, to construct the physical 3D geometry from these 2D observations, the ``Build'' phase is seamlessly interleaved within this active loop. At its core lies a novel \textbf{Semantic-Anchored Geometric Expansion} mechanism, explicitly designed to overcome the coverage deficits inherent to brittle VLM tracking. By mathematically projecting a semantically anchored 3D centroid across the entire video sequence, this mechanism robustly acquires complete multi-view 2D masks. The agent then utilizes continuous geometric tool invocations to lift these masks directly into 3D space, actively constructing a structurally complete point cloud and estimating a precise 3D bounding box.

\vspace{-10pt}

In summary, our main contributions are: (1) We propose \textbf{\textit{TAB}}, a novel agentic framework that reformulates zero-shot 3D visual grounding as an active semantic reasoning and geometric reconstruction process. (2) We introduce a Semantic-Anchored Geometric Expansion mechanism to overcome the multi-view coverage deficit inherent in purely semantic-driven tracking. (3) We identify and correct critical flaws in existing benchmarks (e.g., ScanRefer \citep{chen2020scanrefer} and Nr3D \citep{achlioptas2020referit3d}) to ensure rigorous assessment for future research. (4) Extensive experiments demonstrate that \textbf{\textit{TAB}}, powered entirely by open-source models, significantly outperforms previous zero-shot methods and even surpasses fully supervised baselines in localizing 3D targets.

\vspace{-5pt}
\section{Related Works}
\vspace{-5pt}

\textbf{3D Visual Grounding} aims to precisely localize a target object within a 3D scene based on a natural language query. Historically, the field has been dominated by fully supervised methods, primarily divided into two-stage pipelines \citep{d3net2022d, jain2022bottom} that rely on pre-trained 3D detectors to generate proposals, and single-stage architectures \citep{qian2024multi, wu2023eda} that directly fuse point cloud and textual features. While achieving strong performance, these approaches are bottlenecked by the prohibitive cost of dense 3D annotations and their limited generalization to open-vocabulary, real-world scenarios. To circumvent this, recent pioneering zero-shot methods, such as LLM-Grounder \citep{Yang2023LLMGrounder}, SPAZER \citep{jin2025spazer}, and SeeGround \citep{li2025seeground}, have leveraged the remarkable reasoning capabilities of LLMs and VLMs. However, these training-free methods typically employ static workflows and heavily depend on pre-scanned 3D point clouds and 3D detectors, essentially degrading the grounding process into a discrete classification task of existing proposals. Among proposal-free approaches, VLM-Grounder \citep{xu2025vlmgrounder} is the most closely related to TAB because it also performs grounding from multi-view RGB-D observations without requiring 3D proposals. However, VLM-Grounder primarily performs semantic association across views before aggregating the matched observations in 3D, making them highly vulnerable to occlusions and extreme camera angles where semantic features degrade. TAB instead uses semantic reasoning to establish an initial 3D target anchor and subsequently projects this anchor back into previously unobserved frames to recover additional masks. Thus, geometry in TAB actively guides evidence acquisition rather than serving only as a final aggregation mechanism.

\textbf{VLMs for 3D Understanding.} The remarkable success of 2D Vision-Language Models (VLMs) \citep{li2024llavaonevision, Qwen3-VL, xu2025slowfast1.5, wang2025streambridge} has spurred extensive efforts to extend their perception capabilities into 3D environments. Current 3D Large Multimodal Models generally construct 3D-aware representations by either employing specialized 3D encoders to process point clouds directly (e.g., PointLLM \citep{xu2024pointllm}, SpatialLM \citep{mao2025spatiallm}), or by aggregating multi-view 2D image features into unified 3D tokens (e.g., 3D-LLM \citep{3dllm}, LLaVA-3D \citep{zhu2025llava3D}, Video-3D LLM\citep{zheng2025video3dllm}). While these models demonstrate impressive scene-level reasoning and dialogue capabilities, they rely heavily on massive datasets of paired 3D-text annotations for pretraining. Furthermore, their architecture strictly requires explicit, dense 3D inputs during inference, such as pre-reconstructed point clouds or voxel grids. In contrast, our \textbf{\textit{TAB}} framework circumvents the need for 3D-specific pre-training. We harness the inherent reasoning of 2D VLMs within an agentic loop. By orchestrating foundation models with geometric projections, our approach achieves precise 3D spatial understanding from raw video streams.

\vspace{-5pt}
\section{Method}
\vspace{-5pt}

\begin{figure*}
    \centering
    \includegraphics[width=1\linewidth]{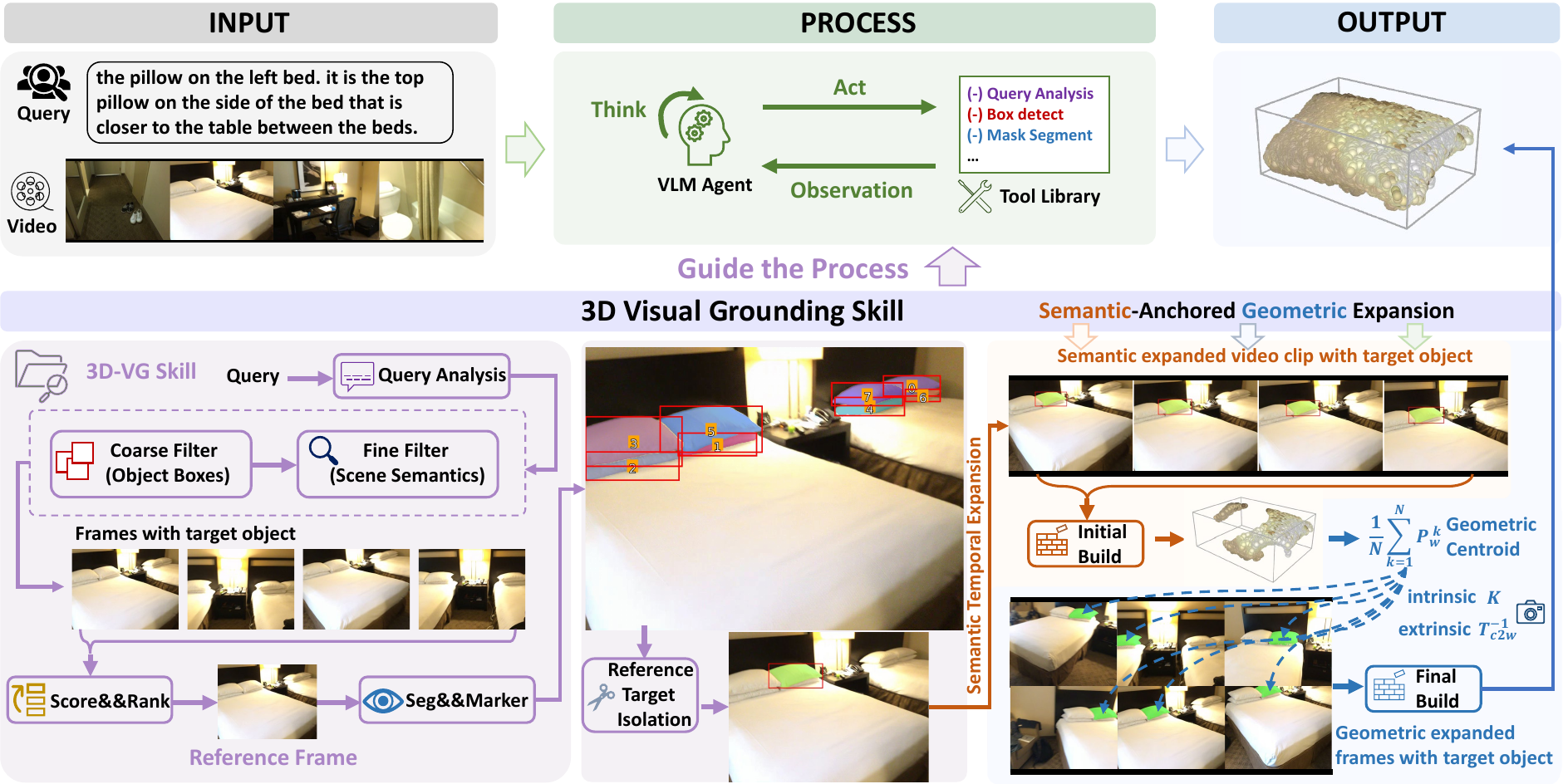}
    \caption{The "\textbf{\textit{Think, Act, Build (TAB)}}" framework. Guided by an expert 3D-VG Skill, the VLM agent reasons about the task, invokes visual tools, and reconstructs the target.}
    \label{fig:method}
    \vspace{-10pt}
\end{figure*}

Given a language query $\mathcal{Q}$ and sequential RGB-D video streams $\mathcal{V} = \{(I_i, D_i)\}_{i=1}^T$ consisting of $T$ frames (where $I_i$ is the RGB image and $D_i$ is the aligned depth map) with camera intrinsics $\mathbf{K}$ and extrinsics $\mathbf{T}_{c2w}$, \textbf{\textit{TAB}} directly reconstructs the target object and calculates the 3D bounding box $\mathbf{B} \in \mathbb{R}^6$. As in Figure \ref{fig:method}, \textbf{\textit{TAB}} formulates 3D-VG as a dynamic agentic loop governed by a comprehensive \textbf{3D-VG Skill} blueprint. This expert skill serves as the master execution plan, dictating how the VLM agent iteratively engages in a \textbf{Think} (contextual reasoning and planning) and \textbf{Act} (invoking specialized tools) paradigm. Crucially, the \textbf{Build} phase is seamlessly interleaved within this loop to overcome the brittleness of purely semantic tracking.

\subsection{Reference Target Localization}

The agent localizes the initial reference target by orchestrating a suite of visual tools, guided by its reasoning thoughts from the 3D-VG Skill.

\textbf{Query Analysis.} In the initial action, the agent invokes the \textit{Query Analysis} tool to parse the raw, free-form text $\mathcal{Q}$ into a structured JSON format. A complex query like "the pillow on the left bed..." is explicitly disentangled into a target class ("pillow"), visual attributes ("top pillow"), spatial conditions ("closer to the table"), and global scene features ("between the beds"), which are then routed to downstream tools as execution arguments.

\textbf{Coarse-to-Fine Filtering.} To locate candidate frames, the agent executes a two-stage filtering. First, the \textit{Coarse Filter} tool utilizes foundation detectors to retain frames containing the target class (e.g., identifying frames with beds and pillows). Because conventional detectors fail to distinguish the specific queried instance from same-class distractors, the \textit{Fine Filter} tool prompts the VLM to rigorously verify if the remaining frames satisfy the parsed scene constraints (e.g., ensuring the frame actually depicts "the table between the beds").

\textbf{Target Isolation.} To transition from frame-level retrieval to instance-level grounding, the agent invokes the \textit{Score}\&\textit{Rank} tool to evaluate the candidate frames against the query's parsed attributes and conditions, selecting the most highly-scored image as the \textit{Reference Frame}. Since multiple instances of the target class may co-exist within a single scene, the agent must resolve intra-class ambiguity. To achieve this, the \textit{Seg}\&\textit{Marker} tool utilizes segmentation models (e.g., SAM3 \citep{carion2026sam3}) to segment all objects of the target class in the reference frame, overlaying a unique ID on each mask. During the subsequent \textit{Reference Target Isolation} step, the VLM actively reasons over this visually prompted image alongside the parsed query to deduce the exact target ID (e.g., singling out the specific top pillow). The agent then isolates and only retains the mask associated with this specific ID. At this stage, the agent has successfully anchored the unambiguous target instance, establishing a pristine visual reference $I_{ref}$.

\subsection{Semantic-Anchored Geometric Expansion} 
\label{subsec:expansion}

While VLMs exhibit profound reasoning, pure 2D semantic tracking is brittle. As shown by the ``pillow'' query in Figure \ref{fig:method}, textual semantics (e.g., ``closer to the table'') often fail under extreme viewpoint variations or close-up views lacking context, leading to a multi-view coverage deficit. To overcome this, we introduce the \textbf{Semantic-Anchored Geometric Expansion} mechanism, which executes a strategic $2D \rightarrow 3D \rightarrow 2D$ mapping. First, through \textit{Semantic Temporal Expansion}, the agent tracks the target locally to construct an \textit{Initial Build} ($2D \rightarrow 3D$) and extract a stable 3D geometric centroid ($\mathbf{P}_{centroid}$). Next, through \textit{Multi-View Geometric Expansion}, the agent mathematically projects this $\mathbf{P}_{centroid}$ onto globally unobserved frames ($3D \rightarrow 2D$), leveraging deterministic geometry to acquire complete multi-view masks and bypass VLM semantic blind spots.

\begin{figure}[t]
  \hrule height 0.8pt \vspace{0.5ex}
  \captionof{algorithm}{Semantic Temporal Expansion}
  \label{alg:expansion}
  \vspace{-9pt} \hrule height 0.4pt \vspace{0.5ex}
  \begin{algorithmic}[1]
    \Require $\mathcal{V} = \{(I_i, D_i)\}_{i=1}^T, \mathcal{Q}, (I_{ref}, D_{ref}, M_{ref})$
    \Ensure $\mathcal{V}_{sem}$
    
    \State $\mathcal{V}_{sem} \gets \{(I_{ref}, D_{ref}, M_{ref})\}$
    \For{\textbf{direction} $\Delta t \in \{+1, -1\}$}
        \State $t \gets t_{ref} + \Delta t$
        \While{$1 \le t \le T$}
            \If{\text{VLM\_Verify}($\mathcal{V}_{sem}, I_t, \mathcal{Q}$)}
                \State $M_t \gets \text{Segmentation}(I_t)$
                \State $\mathcal{V}_{sem} \gets \mathcal{V}_{sem} \cup \{(I_t, D_t, M_t)\}$
                \State $t \gets t + \Delta t$
            \Else \State \textbf{break}
            \EndIf
        \EndWhile
    \EndFor
  \end{algorithmic}
  \vspace{0.5ex} \hrule height 0.8pt
  \vspace{-10pt}
\end{figure} 

\textbf{Semantic Temporal Expansion.} Relying on a single 2D reference frame to lift a 3D centroid is highly susceptible to noise and self-occlusion. As illustrated in the ``Semantic Temporal Expansion'' arrow of Figure \ref{fig:method}, a short-term, multi-frame context is imperative to construct an initial 3D geometry. In Alg \ref{alg:expansion}, given the single reference frame $I_{ref}$ at temporal index $t_{ref}$, the agent exploits the inherent spatiotemporal continuity of the video stream. It initiates a bidirectional tracking loop along the temporal axis. The agent maintains a dynamically growing video context memory $\mathcal{V}_{sem}$, initialized with the reference frame and its target mask $(I_{ref}, D_{ref}, M_{ref})$. For each adjacent candidate frame $I_t$ expanding outwards from $t_{ref}$, the VLM verifies the object's identity consistency against the context in $\mathcal{V}_{sem}$. If verified, the segmentation model generates the mask $M_t$, and the tuple $(I_t, D_t, M_t)$ is appended to $\mathcal{V}_{sem}$ as the updated context for the next frame. This expansion loop iteratively advances and terminates immediately in a given direction the moment the VLM deduces that the target is no longer present (e.g., due to severe occlusion or moving out of the view). This semantic-driven tracking captures the object across slightly varying local viewpoints, yielding a continuous and highly reliable semantic video clip $\mathcal{V}_{sem} = \{(I_t, D_t, M_t)\}_{t \in \mathcal{T}_{local}}$, where $\mathcal{T}_{local}$ is the set of successfully tracked frame indices.

\textbf{Centroid Extraction.} When tracking frames with the target object, strictly enforcing spatial conditions (e.g., ``next to the piano'') severely limits multi-view coverage, as VLMs can mistakenly reject valid frames with close-up or different views of the target object but lacking these contextual backgrounds. To solve this, we abstract the locally tracked 2D pixels into an immutable, viewpoint-invariant 3D physical anchor. Specifically, the agent first reconstructs the target by inverse-projecting \textit{only} the foreground pixels enclosed by the target mask (i.e., $(u, v) \in M_t$) from each frame $I_t$ in $\mathcal{V}_{sem}$ into 3D points. Assuming a pinhole camera model, we utilize the true physical depth $D_t(u, v)$ provided by the aligned depth map, alongside the camera intrinsic matrix $\mathbf{K} \in \mathbb{R}^{3 \times 3}$ to recover the corresponding 3D point $\mathbf{P}_c = [x_c, y_c, z_c]^T$ from the 2D pixel $(u, v)$ in the local camera coordinate system:
\begin{equation}
    \mathbf{P}_c = D_t(u, v) \cdot \mathbf{K}^{-1} \begin{bmatrix} u \\ v \\ 1 \end{bmatrix}, \mathbf{K} = \begin{bmatrix} f_x & 0 & c_x \\ 0 & f_y & c_y \\ 0 & 0 & 1 \end{bmatrix}
    \label{eq:pc}
\end{equation}
where $f_x, f_y$ represent the focal lengths of the camera, and $c_x, c_y$ denote the principal point. To aggregate these object points into a globally consistent environment, $\mathbf{P}_c$ must also be transformed into the absolute 3D world coordinate system $\mathbf{P}_w = [x_w, y_w, z_w]^T$. Using the frame-specific camera extrinsic $\mathbf{T}_{c2w} \in \mathbb{R}^{4 \times 4}$, which encapsulates the rotation $\mathbf{R} \in \mathbb{R}^{3 \times 3}$ and translation $\mathbf{t} \in \mathbb{R}^{3}$, the transformation is formulated as:
\begin{equation}
    \begin{bmatrix} \mathbf{P}_w \\ 1 \end{bmatrix} = \mathbf{T}_{c2w} \begin{bmatrix} \mathbf{P}_c \\ 1 \end{bmatrix} = \begin{bmatrix} \mathbf{R} & \mathbf{t} \\ \mathbf{0}^T & 1 \end{bmatrix} \begin{bmatrix} \mathbf{P}_c \\ 1 \end{bmatrix}
    \label{eq:pw}
\end{equation}
Since the inverse-projection is strictly confined to pixels in the masked regions, aggregating these lifted world points $\mathbf{P}_w$ across $\mathcal{V}_{sem}$ naturally yields a preliminary 3D point cloud for the target object, denoted as the \textit{Initial Build} ($PCD_{init}$). From this isolated 3D structure, the framework calculates the physical geometric centroid $\mathbf{P}_{centroid} \in \mathbb{R}^3$:
\begin{equation}
    \mathbf{P}_{centroid} = \frac{1}{N} \sum_{k=1}^{N} \mathbf{P}_{w}^k
\end{equation}
where $\mathbf{P}_{w}^k$ is the $k$-th valid point in $PCD_{init}$, and $N$ is the total number of valid points. This $\mathbf{P}_{centroid}$ serves as the spatial anchor for the subsequent multi-view geometric expansion.

\textbf{Multi-View Geometric Expansion.} 
With the absolute 3D $\mathbf{P}_{centroid}$, the agent can geometrically decide whether a given frame $I_i$ contains the target object, bypassing the VLM's semantic tracking failures. The process follows three key steps: First, the agent mathematically projects the 3D $\mathbf{P}_{centroid}$ back onto the 2D image plane of frame $I_i$. By applying the inverse extrinsic matrix $\mathbf{T}_{c2w}^{-1} \in \mathbb{R}^{4 \times 4}$ and the intrinsic matrix $\mathbf{K}$, we obtain the theoretical 2D pixel coordinates $[u, v]^T$ and the predicted depth $z_{pred}$ of $\mathbf{P}_{centroid}$ in frame $I_i$:
\begin{equation}
    \begin{bmatrix} \mathbf{P}_{c} \\ 1 \end{bmatrix} = \mathbf{T}_{c2w}^{-1} \begin{bmatrix} \mathbf{P}_{centroid} \\ 1 \end{bmatrix}, z_{pred} \begin{bmatrix} u \\ v \\ 1 \end{bmatrix} = \mathbf{K} \mathbf{P}_{c}
\end{equation}

Second, because this direct projection ignores the Field of View (FoV) boundaries and physical occlusions, the agent performs a strict visibility check on $(u, v)$ and $z_{pred}$ to ensure the target is actually observable in frame $I_i$:
\begin{equation}
    \underbrace{(u, v) \in \Omega}_{\substack{\text{FoV Boundary} \\ \text{Check}}} \land 
    \underbrace{z_{actual} > 0}_{\substack{\text{Depth Validity} \\ \text{Check}}} \land 
    \underbrace{z_{pred} \le z_{actual} + \epsilon}_{\substack{\text{Z-buffer Occlusion} \\ \text{Check}}}
\end{equation}
where $\Omega = [0, W) \times [0, H)$ represents the 2D image domain, $z_{actual} = D_i(u, v)$ is the physical depth sampled from the aligned depth map, and $\epsilon$ (set to 0.4) accommodates sensor noise and inherent object thickness. Finally, for each verified frame $I_i$ that passes this check, the 2D coordinate $(u, v)$ serves as a precise point prompt for the segmentation model (e.g., SAM3) to extract the target mask $M_i$. The comprehensive tuple $(I_i, D_i, M_i)$ is then appended to the expansion pool: $\mathcal{V}_{geo} \gets \mathcal{V}_{geo} \cup \{(I_i, D_i, M_i)\}$, deterministically acquiring complete multi-view observations ready for the final dense 3D reconstruction.

\definecolor{darkblue}{rgb}{0, 0, 0.5}
\definecolor{Gray}{gray}{0.93}
\definecolor{uclagold}{rgb}{1.0, 0.7, 0.0}
\definecolor{airforceblue}{rgb}{0.36, 0.54, 0.66}
\definecolor{rosegold}{rgb}{0.72, 0.43, 0.47}
\definecolor{pastelbrown}{rgb}{0.51, 0.41, 0.33}
\definecolor{isabelline}{rgb}{0.96, 0.94, 0.93}
\definecolor{macaroniandcheese}{rgb}{0.98, 0.89, 0.83}
\definecolor{wildblueyonder}{rgb}{0.85, 0.89, 0.95}
\definecolor{mediumtaupe}{rgb}{0.4, 0.3, 0.28}
\definecolor{bluegray}{rgb}{0.4, 0.6, 0.8}
\definecolor{celestialblue}{rgb}{0.29, 0.59, 0.82}
\definecolor{darkorange}{rgb}{1.0, 0.55, 0.0}
\definecolor{cadmiumred}{rgb}{0.89, 0.0, 0.13}
\definecolor{magnolia}{rgb}{0.97, 0.96, 1.0}
\definecolor{pastelblue}{rgb}{0.68, 0.78, 0.81}
\definecolor{persiangreen}{rgb}{0.0, 0.65, 0.58}
\definecolor{steelblue}{rgb}{0.27, 0.51, 0.71}
\definecolor{bluebell}{rgb}{0.64, 0.64, 0.82}
\definecolor{dimgray}{rgb}{0.41, 0.41, 0.41}
\definecolor{splashedwhite}{rgb}{1.0, 0.99, 1.0}
\definecolor{lavendergray}{rgb}{0.77, 0.76, 0.82}
\definecolor{lightgray}{rgb}{0.83, 0.83, 0.83}
\definecolor{lavendermist}{rgb}{0.9, 0.9, 0.98}
\definecolor{lightgreen}{HTML}{f8fcf4}
\definecolor{lightblue}{HTML}{dfebf7}

\definecolor{peachpuff}{HTML}{FFDAB9}      
\definecolor{mistyrose}{HTML}{FFE4E1}      
\definecolor{lavenderblush}{HTML}{FFF0F5}  
\definecolor{lemonchiffon}{HTML}{FFFACD}   
\definecolor{cornsilk}{HTML}{FFF8DC}       
\definecolor{seashell}{HTML}{FFF5EE}       

\definecolor{aliceblue}{HTML}{F0F8FF}      
\definecolor{paleblue}{rgb}{0.85, 0.92, 0.98} 
\definecolor{lightcyan}{HTML}{E0FFFF}      
\definecolor{mintcream}{HTML}{F5FFFA}      
\definecolor{honeydew}{HTML}{F0FFF0}       
\definecolor{palegreen}{rgb}{0.85, 0.96, 0.85} 

\definecolor{floralwhite}{HTML}{FFFAF0}    
\definecolor{whitesmoke}{HTML}{F5F5F5}     
\definecolor{palepurple}{rgb}{0.92, 0.88, 0.95} 
\definecolor{oatmeal}{rgb}{0.94, 0.92, 0.88}    

\definecolor{palelavender}{HTML}{F5F2F9} 
\definecolor{lilacwhite}{HTML}{F8F6FA}   
\definecolor{ghostpurple}{rgb}{0.96, 0.95, 0.98} 
\definecolor{softthistle}{rgb}{0.93, 0.91, 0.95}

\begin{table*}[!t]\Huge
    \centering
    \resizebox{\linewidth}{!}{
    \begin{tabular}{c | c | c | c c | c c | c c}
        \toprule
        \multirow{2}{*}{\textbf{Method}} & \multirow{2}{*}{\textbf{LLM/VLM}} & \multicolumn{1}{c|}{\textbf{w/}} & \multicolumn{2}{c|}{\textbf{Unique}} & \multicolumn{2}{c|}{\textbf{Multiple}} & \multicolumn{2}{c}{\textbf{Overall}} \\

        \cmidrule(lr){4-5} \cmidrule(lr){6-7} \cmidrule(lr){8-9}
        
         &  &\textbf{PC.} &\textbf{Acc@0.25} &\textbf{Acc@0.5} &\textbf{Acc@0.25} &\textbf{Acc@0.5} &\textbf{Acc@0.25} &\textbf{Acc@0.5} \\

        \midrule
        \rowcolor{aliceblue} 
        \multicolumn{9}{c}{\textbf{Fully-Supervised Methods (One/Two-Stage Based)}}\\
        \midrule
        ScanRefer \citep{chen2020scanrefer}     & - &\cmark & 67.6 & 46.2 & 32.1 & 21.3 & 39.0 & 26.1 \\
        3DVG-T \citep{zhao20213dvg}             & - &\cmark & 77.2 & 58.5 & 38.4 & 28.7 & 45.9 & 34.5 \\
        BUTD-DETR \citep{jain2022bottom}        & - &\cmark & 84.2 & 66.3 & 46.6 & 35.1 & 52.2 & 39.8 \\
        EDA \citep{wu2023eda}                   & - &\cmark & 85.8 & 68.6 & 49.1 & 37.6 & 54.6 & 42.3 \\
        G3-LQ \citep{wang2024g3lq}              & - &\cmark & 88.6 & 73.3 & 50.2 & 39.7 & 56.0 & 44.7 \\

        \midrule
        \rowcolor{seashell} 
        \multicolumn{9}{c}{\textbf{Fully-Supervised Methods (LLM/VLM Based)}}\\
        \midrule
        LLaVA-3D \citep{zhu2025llava3D}         & LLaVA-Video-7B   &\xmark & -   & -   & -   & -   &50.1 &42.7 \\
        3D-LLaVA \citep{deng20253dllava}        & LLaVA-1.5-7B     &\cmark & -   & -   & -   & -   &51.2 &40.6 \\
        Chat-Scene \citep{huang2024chatscene}   & Vicuna-7B        &\cmark &89.5 &82.4 &47.7 &42.9 &55.5 &50.2 \\
        SPAR-mix \citep{zhang2025spar}           & InternVL-2.5-8B &\xmark 
        & -   & -   & -   & -   &31.9 &12.4 \\
        VG-LLM \citep{zheng2025vgllm}           & Qwen2.5-VL-7B    &\xmark  & -   & -   & -   & -   &41.6 &14.9 \\
        Video-3D-LLM \citep{zheng2025video3dllm}& LLaVA-Video-7B   &\cmark &87.9 &78.3 &50.9 &45.3 &58.1 &51.7 \\
        GPT4Scene \citep{qi2026gpt4scene}       & Qwen2-VL-7B      &\cmark &90.3 &83.7 &56.4 &50.9 &62.6 &57.0 \\
        3D-RS \citep{huang20253drs}             & LLaVA-Next-Video &\cmark &87.4 &77.9 &57.0 &50.8 &62.9 &56.1 \\

        \midrule
        \rowcolor{lightgreen} 
        \multicolumn{9}{c}{\textbf{Zero-Shot Methods}}\\
        \midrule
        
        LLM-Grounder \citep{Yang2023LLMGrounder} & GPT-4 turbo          &\cmark  & -   & -   & -   & -   &17.1 &5.3\\
        ZSVG3D \citep{yuan2024zsvg3d}            & GPT-4 turbo          &\cmark  &63.8 &58.4 &27.7 &24.6 &36.4 &32.7\\
        SeeGround \citep{li2025seeground}        & Qwen2-VL-72B         &\cmark  &75.7 &68.9 &34.0 &30.0 &44.1 &39.4\\
        CSVG  \citep{yuan2024csvg}               & Mistral-Large-2407   &\cmark  &68.8 &61.2 &38.4 &27.3 &49.6 &39.8\\
        VLM-Grounder \citep{xu2025vlmgrounder}   & GPT-4o               &\xmark  &66.0 &29.8 &48.3 &33.5 &51.6 &32.8\\
        SeqVLM \citep{lin2025seqvlm}             & Doubao-1.5-pro       &\cmark  &77.3 &72.7 &47.8 &41.3 &55.6 &49.6\\
        SPAZER \citep{jin2025spazer}             & GPT-4o               &\cmark  &80.9 &72.3 &51.7 &43.4 &57.2 &48.8\\
        
        \midrule
        \rowcolor{lightgray!30}
        \textbf{\textit{TAB}} \textbf{(ours)}    & Qwen3-VL-32B         &\xmark  &90.2 &57.6 &60.1 &39.9 &71.2 &46.4 \\
        \rowcolor{lightgray!30}
        \textbf{\textit{TAB}} \textbf{(ours)}    & Qwen3-VL-32B         &\cmark  &90.2 &77.2 &60.8 &52.5 &71.6 &61.6\\

        \bottomrule
    \end{tabular}
}
\caption{3D Visual Grounding results on ScanRefer. \textbf{"w/ PC"} denotes reliance on 3D Point Clouds and Proposals.}
\label{tab:scanref}
\vspace{-5pt}
\end{table*}

\subsection{2D to 3D Reconstruction}
Given the multi-view observations $\mathcal{V}_{geo}$, \textbf{\textit{TAB}} inverse-projects the masked pixels in $\mathcal{V}_{geo}$ into the 3D world coordinate system following the same process as in Eq. (\ref{eq:pc}) and Eq. (\ref{eq:pw}). To mitigate depth sensor noise and segmentation artifacts, we filter the aggregated raw point cloud using Statistical Outlier Removal \citep{rusu2010sor} and DBSCAN clustering \citep{ester1996dbscan} to isolate the main object geometry. Finally, we compute the spatial extremes of this clean cluster to estimate the axis-aligned 3D bounding box $\mathbf{B} \in \mathbb{R}^6$, successfully completing the zero-shot 3D grounding process without relying on pre-scanned point clouds. Moreover, unlike static zero-shot pipelines that fail upon a single intermediate error, our \textbf{\textit{TAB}} features robust fault tolerance. By actively monitoring its iterative progress, the agent autonomously recovers from local failures. For example, if a filtering step yields zero candidate images or depth noise corrupts the initial build, it employs a ``Dynamic Adjustment'' strategy to proactively relax tool thresholds or just skip non-critical steps. This adaptability ensures continuous execution and successful 3D grounding in noisy environments.

\vspace{-5pt}
\section{Benchmark Refinement}
\label{sec:refinement}
While benchmarks such as ScanRefer \cite{chen2020scanrefer} and Nr3D \cite{achlioptas2020referit3d} have advanced 3D-VG, we observe non-negligible annotation noise within the widely adopted evaluation subsets from \citep{xu2025vlmgrounder}. To ensure a rigorous assessment, we manually reviewed and refined these annotations, categorizing the errors into three primary types, including \textbf{Ambiguous References}, \textbf{Object Category Errors}, and \textbf{Spatial Location Errors} as in Appendix Figure \ref{fig:refine}.

\begin{table*}[!t]\footnotesize 
    \centering
    \begin{tabular}{c | c | c | c | c | c | c | c}
        \toprule
        \textbf{Method} & \textbf{LLM/VLM} &\textbf{w/ PC.} &\textbf{Easy} & \textbf{Hard} & \textbf{Dep.} & \textbf{Indep.} & \textbf{Overall} \\
        
        \midrule
        \rowcolor{aliceblue} 
        \multicolumn{8}{c}{\textbf{Fully-Supervised Methods}}\\
        \midrule
         ReferIt3DNet \citep{achlioptas2020referit3d} &- &\cmark &43.6 &27.9 &32.5 &37.1 &35.6\\
         InstanceRefer \citep{yuan2021instancerefer}  &- &\cmark &46.0 &31.8 &34.5 &41.9 &38.8\\
        3DVG-T \citep{zhao20213dvg}                  &- &\cmark &48.5 &34.8 &34.8 &43.7 &40.8\\
        EDA \citep{wu2023eda}                        &- &\cmark &58.2 &46.1 &50.2 &53.1 &52.1\\
        BUTD-DETR \citep{jain2022bottom}             &- &\cmark &60.7 &48.4 &46.0 &58.0 &54.6\\
        3D-VisTA \citep{zhu20233dvista}              &- &\cmark &72.1 &56.7 &61.5 &65.1 &64.2\\
        SceneVerse \citep{jia2024sceneverse}         &- &\cmark &72.5 &57.8 &56.9 &67.9 &64.9\\

        \midrule
        \rowcolor{lightgreen} 
        \multicolumn{8}{c}{\textbf{Zero-Shot Methods}}\\
        \midrule
        ZSVG3D \citep{yuan2024zsvg3d}            & GPT-4 turbo    &\cmark &46.5 &31.7 &36.8 &40.0 &39.0\\
        SeeGround \citep{li2025seeground}        & Qwen2-VL-72B   &\cmark &54.5 &38.3 &42.3 &48.2 &46.1\\
        VLM-Grounder \citep{xu2025vlmgrounder}   & GPT-4o         &\xmark &55.2 &39.5 &45.8 &49.4 &48.0\\
        SeqVLM \citep{lin2025seqvlm}             & Doubao-1.5-pro &\cmark &58.1 &47.4 &51.0 &54.5 &53.2\\
        SPAZER \citep{jin2025spazer}             & GPT-4o         &\cmark &68.0 &58.8 &59.9 &66.2 &63.8\\

        \midrule
        \rowcolor{lightgray!30}
        \textbf{\textit{TAB}} \textbf{(ours)}    & Qwen3-VL-32B   &\xmark &72.1 &63.2 &62.5 &71.4 &68.0 \\  
        \rowcolor{lightgray!30}
        \textbf{\textit{TAB}} \textbf{(ours)}    & Qwen3-VL-32B   &\cmark &72.1 &61.4 &63.5 &69.5 &67.2\\     
        \bottomrule
    \end{tabular}

\caption{3D Visual Grounding results on Nr3D. \textbf{"w/ PC"} denotes reliance on 3D Point Clouds and Proposals.}
\label{tab:nr3d}
\vspace{-5pt}
\end{table*}

\section{Experiments}
\label{sec:experiment}
\textbf{Benchmarks and Metrics.} 
We experiment on the ScanRefer \citep{chen2020scanrefer} and Nr3D \citep{achlioptas2020referit3d} benchmarks, both built upon ScanNet \citep{dai2017scannet} indoor scenes. ScanRefer queries are categorized as ``Unique'' or ``Multiple'' depending on the presence of same-class distractors. Its performance is measured by \textbf{Acc@0.25} and \textbf{Acc@0.5}, representing the fraction of predicted 3D bounding boxes with an IoU $>0.25$ and $0.5$ against the ground truth. Nr3D queries are divided into ``Easy''/``Hard'' and ``View-Dependent''/``Independent'' subsets. Its performance is evaluated by top-1 selection accuracy. Our main evaluations are conducted on the widely adopted subsets following previous works \citep{xu2025vlmgrounder, jin2025spazer, lin2025seqvlm}.

\textbf{Evaluation Settings.} We present both our native predictions operating solely on raw RGB-D streams (\textbf{w/ PC.} = \xmark) and an assisted setting refined by Mask3D proposals (\textbf{w/ PC.} = \cmark). This assisted setting is not required by our framework, but is provided exclusively for equitable comparisons with prior methods (e.g., SeqVLM, SPAZER, SeeGround) that inherently rely on off-the-shelf 3D proposals. No ground-truth boxes or ground-truth IoU are used in this setting. We select the Mask3D proposal with the highest 3D IoU against TAB's proposal-free reconstructed box.

\textbf{Implementation Details.} We build our framework with the Qwen3-VL-32B \citep{Qwen3-VL} as the primary VLM agent. For the foundational vision tools, we utilize Grounding DINO \citep{liu2024groundingdino} for coarse object detection and SAM3 \citep{carion2026sam3} for instance segmentation. Both the Semantic Temporal Expansion and the Multi-View Geometric Expansion processes are capped at a maximum of $32$ frames. 

\subsection{3D Visual Grounding Results}

\textbf{ScanRefer.} For rigorous comparison, we decouple the evaluation into \textbf{Proposal-Free (w/ PC. = \xmark)} and \textbf{Proposal-Assisted (w/ PC. = \cmark)} settings (Table \ref{tab:scanref}). Operating strictly on raw RGB-D streams, \textbf{\textit{TAB}} achieves an Acc@0.25 of 71.2\% and Acc@0.5 of 46.4\%, outperforming VLM-Grounder, a similarly proposal-free method, by a massive absolute margin of +19.6\% and +13.6\%. To establish comparison with methods inherently reliant on 3D priors, following previous works \citep{lin2025seqvlm, jin2025spazer, li2025seeground}, our \textbf{Proposal-Assisted} setting refines \textbf{\textit{TAB}}'s natively reconstructed boxes using the Mask3D \citep{schult23mask3d} generated proposal that yields the maximum 3D IoU overlap. Crucially, this incorporation yields a substantial surge in Acc@0.5 (46.4\% $\rightarrow$ 61.6\%), dominating all prior zero-shot and fully-supervised methods by a large margin, while Acc@0.25 remains remarkably stable (71.2\% $\rightarrow$ 71.6\%). This stability is a key indicator: it proves that \textbf{\textit{TAB}} intrinsically localizes the correct targets entirely on its own, and external 3D proposals merely act as a geometric refinement to tighten boundary precision. Ultimately, even in the strict proposal-free setting, \textbf{\textit{TAB}} still largely surpasses previous proposal-assisted methods in Acc@0.25 (e.g., 71.2\% vs SPAZER's 57.2\%) and remains highly competitive in Acc@0.5. More importantly, our framework achieves this while completely bypassing the need for pre-scanned 3D point clouds and off-the-shelf 3D detectors, a critical capability for real-world deployment that proposal-dependent pipelines fundamentally cannot replicate.


\textbf{Nr3D.} The evaluation protocol of Nr3D (top-1 selection) perfectly isolates grounding capability from bounding box regression noise. As in Table \ref{tab:nr3d}, operating under the \textbf{Proposal-Free (w/ PC = \xmark)} setting, \textbf{\textit{TAB}} establishes a new zero-shot state-of-the-art with an overall accuracy of 68.0\%. Despite the absence of 3D Scene point clouds, \textbf{\textit{TAB}} significantly outperforms prior zero-shot approaches that utilize these 3D priors (e.g., SPAZER at 63.8\%) and even surpasses fully-supervised baselines like SceneVerse (64.9\%). The framework's robustness is particularly evident in the challenging "Hard" and "View-Dependent" subsets, achieving accuracies of 63.2\% and 62.5\%, respectively. By actively reasoning over visual semantics to disambiguate complex multi-object references and utilizing multi-view geometry to bypass semantic tracking blind spots, \textbf{\textit{TAB}} overcomes the structural limitations that typically paralyze static pipelines.

\begin{table*}[!t]\footnotesize
\centering
\resizebox{\linewidth}{!}{
\begin{tabular}{c | cc | cc cc cc | ccccc}
\toprule
\multirow{3}{*}{\textbf{\#}} & \multicolumn{2}{c|}{\textbf{Modules}} & \multicolumn{6}{c|}{\textbf{ScanRefer}} & \multicolumn{5}{c}{\textbf{Nr3D}} \\
\cmidrule(lr){2-3} \cmidrule(lr){4-9} \cmidrule(lr){10-14}
& \multirow{2}{*}{\textbf{STE}} & \multirow{2}{*}{\textbf{MGE}} & \multicolumn{2}{c}{\textbf{Unique}} & \multicolumn{2}{c}{\textbf{Multiple}} & \multicolumn{2}{c|}{\textbf{Overall}} & \multirow{2}{*}{\textbf{Easy}} & \multirow{2}{*}{\textbf{Hard}} & \multirow{2}{*}{\textbf{Dep.}} & \multirow{2}{*}{\textbf{Indep.}} & \multirow{2}{*}{\textbf{Overall}} \\
\cmidrule(lr){4-5} \cmidrule(lr){6-7} \cmidrule(lr){8-9}
& & & \textbf{@0.25} & \textbf{@0.5} & \textbf{@0.25} & \textbf{@0.5} & \textbf{@0.25} & \textbf{@0.5} & & & & & \\
\midrule
(a) & \xmark & \xmark &57.6 &29.3 &32.3 &19.0 &41.6 &22.8 &58.8 &43.9 &51.0 &52.6 &52.0 \\
(b) & \xmark & \cmark &65.2 &41.3 &41.1 &22.2 &50.0 &29.2 &61.8 &47.4 &49.0 &59.1 &55.2 \\
(c) & \cmark & \xmark &69.6 &45.7 &51.3 &30.4 &58.0 &36.0 &67.6 &49.1 &61.5 &57.8 &59.2 \\
\rowcolor{lightgray!30}
(d) & \cmark & \cmark &\textbf{90.2} &\textbf{57.6} &\textbf{60.1} &\textbf{39.9} &\textbf{71.2} &\textbf{46.4} &\textbf{72.1} &\textbf{63.2} &\textbf{62.5} &\textbf{71.4} &\textbf{68.0} \\
\bottomrule
\end{tabular}
 }
\caption{Ablation study on the Semantic-Anchored Geometric Expansion. \textbf{STE}: Semantic Temporal Expansion. \textbf{MGE}: Multi-View Geometric Expansion.}
\label{tab:ablation}
\vspace{-10pt}
\end{table*}

\subsection{In-Depth Analysis}

\textbf{Single-Frame Reconstruction.} 
We first establish a naive baseline (Table \ref{tab:ablation} (a)) by disabling both the Semantic Temporal Expansion and the Multi-View Geometric Expansion. The target object is reconstructed only by inverse-projecting the mask from the single isolated reference frame $I_{ref}$. As expected, relying on a single 2D view is highly susceptible to depth sensor noise and severe self-occlusion. This lack of multi-view context prevents the agent from observing the object's full physical extent, resulting in a low overall accuracy of 41.6\% Acc@0.25 on ScanRefer and only 52.0\% on Nr3D.

\textbf{Effect of Semantic Temporal Expansion.} 
To evaluate the necessity of video context, we bypass the STE phase (Table \ref{tab:ablation} (b)) and calculate the initial physical centroid directly from the single reference image $I_{ref}$. This leads to a significant performance drop, particularly in complex scenarios like the Nr3D "Dep." subset (dropping to 49.0\%) and ScanRefer's "Multiple" subset (41.1\% Acc@0.25 and 22.2\% Acc@0.5). Because a single viewpoint only captures a partial surface, the resulting centroid is heavily biased and spatially offset from the object's true center of mass. This inaccurate spatial anchor subsequently corrupts the deterministic geometric projection, causing the visibility checks to fetch misaligned 2D views and ultimately derailing the dense reconstruction. This confirms that exploiting spatiotemporal continuity is imperative for establishing a robust initial 3D geometry.

\textbf{Effect of Multi-View Geometric Expansion.}
We ablate the geometric projection mechanism (Table \ref{tab:ablation} (c)) by reconstructing the target with only the VLM-tracked frames. Tracking driven solely by 2D semantics is inherently brittle. When the camera undergoes extreme viewpoints or occlusions, semantic matching often fails, causing the temporal expansion to terminate. This multi-view coverage deficit is reflected in the sharp decline of localization precision: ScanRefer overall Acc@0.5 drops from 46.4\% to 36.0\%, and performance on the Nr3D "Hard" queries falls from 63.2\% to 49.1\%.

\begin{figure}
    \centering
    \includegraphics[width=1\linewidth]{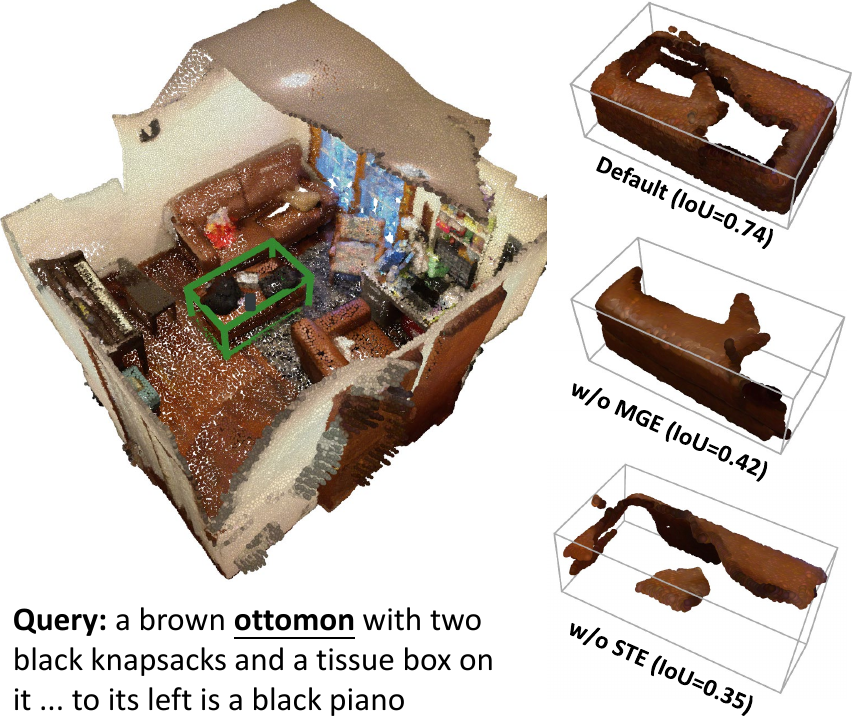}
    \caption{Qualitative comparison of STE and MGE.}
    \label{fig:qualitive}
\vspace{-10pt}
\end{figure}

\textbf{Qualitative Comparison.}
Figure \ref{fig:qualitive} provides a visual comparison of our framework against its ablated variants. By default, \textbf{\textit{TAB}} successfully aggregates complete multi-view observations to predict a tight, highly accurate 3D bounding box (IoU = $0.74$). In contrast, ablating the Multi-View Geometric Expansion (w/o MGE) restricts the agent to pure semantic tracking, which suffers from view coverage deficits, yielding a fragmented geometry and a reduced IoU of $0.42$. Similarly, bypassing the Semantic Temporal Expansion (w/o STE) forces the agent to extract the physical centroid from a single frame. This introduces bias and depth noise into the geometric projection, resulting in an offset and inaccurate bounding box (IoU = $0.35$). These visualizations reinforce the necessity of synergizing both modules for robust 3D localization.

\section{Conclusion}
We present \textbf{\textit{Think, Act, Build (TAB)}}, an agentic framework that reformulates zero-shot 3D Visual Grounding into a dynamic reasoning and reconstruction process. By decoupling semantics from geometry, TAB utilizes 2D Vision-Language Models in an iterative "Think-Act" loop. To overcome semantic tracking limitations, our Semantic-Anchored Geometric Expansion projects a 3D centroid across frames to harvest multi-view masks, seamlessly "Building" complete 3D point clouds from RGB-D streams. Furthermore, we refined existing benchmarks for rigorous evaluation. Ultimately, TAB provides a complementary proposal-free paradigm for zero-shot 3D grounding from sequential RGB-D observations, while improving efficiency and robustness to sensing noise remain important directions for future work.

\section*{Limitations} 
While \textbf{\textit{TAB}} advances zero-shot 3D-VG, it relies on the availability of aligned RGB-D streams, which may limit its application in outdoor environments lacking reliable depth sensors. Additionally, the iterative nature of the ``Think-Act'' agentic loop and the reliance on large foundation models inherently increase inference latency. But we believe these challenges can be mitigated by integrating a stronger depth/camera pose estimation model and inference acceleration engines in the future.

\bibliography{custom}

@String(ACMMM = {ACM MM})

@String(CVPR  = {CVPR})

@String(ICCV  = {ICCV})

@String(ECCV  = {ECCV})

@String(ICLR  = {ICLR})

@String(NIPS  = {NeurIPS})

@String(EMNLP = {EMNLP})

@String(COLM = {COLM})

@String(ICRA = {ICRA})

@String(CORL = {CoRL})

@String(IROS = {IROS})

@inproceedings{chen2020scanrefer,
  title={Scanrefer: 3d object localization in rgb-d scans using natural language},
  author={Chen, Dave Zhenyu and Chang, Angel X and Nie{\ss}ner, Matthias},
  booktitle=ECCV,
  pages={202--221},
  year={2020},
  organization={Springer}
}

@inproceedings{achlioptas2020referit3d,
  title={Referit3d: Neural listeners for fine-grained 3d object identification in real-world scenes},
  author={Achlioptas, Panos and Abdelreheem, Ahmed and Xia, Fei and Elhoseiny, Mohamed and Guibas, Leonidas},
  booktitle=ECCV,
  pages={422--440},
  year={2020},
  organization={Springer}
}

@article{kim2024openvla,
  title={Openvla: An open-source vision-language-action model},
  author={Kim, Moo Jin and Pertsch, Karl and Karamcheti, Siddharth and Xiao, Ted and Balakrishna, Ashwin and Nair, Suraj and Rafailov, Rafael and Foster, Ethan and Lam, Grace and Sanketi, Pannag and others},
  journal={arXiv preprint arXiv:2406.09246},
  year={2024}
}

@inproceedings{anderson2018vln,
  title={Vision-and-language navigation: Interpreting visually-grounded navigation instructions in real environments},
  author={Anderson, Peter and Wu, Qi and Teney, Damien and Bruce, Jake and Johnson, Mark and S{\"u}nderhauf, Niko and Reid, Ian and Gould, Stephen and Van Den Hengel, Anton},
  booktitle={Proceedings of the IEEE conference on computer vision and pattern recognition},
  pages={3674--3683},
  year={2018}
}

@inproceedings{hoenig2015mixed,
  title={Mixed reality for robotics},
  author={Hoenig, Wolfgang and Milanes, Christina and Scaria, Lisa and Phan, Thai and Bolas, Mark and Ayanian, Nora},
  booktitle=IROS,
  pages={5382--5387},
  year={2015},
  organization={IEEE}
}

@inproceedings{roh2022languagerefer,
  title={Languagerefer: Spatial-language model for 3d visual grounding},
  author={Roh, Junha and Desingh, Karthik and Farhadi, Ali and Fox, Dieter},
  booktitle=CORL,
  pages={1046--1056},
  year={2022},
  organization={PMLR}
}

@article{yuan2024csvg,
  title={Solving zero-shot 3d visual grounding as constraint satisfaction problems},
  author={Yuan, Qihao and Li, Kailai and Zhang, Jiaming},
  journal={arXiv preprint arXiv:2411.14594},
  year={2024}
}

@inproceedings{zhao20213dvg,
  title={3dvg-transformer: Relation modeling for visual grounding on point clouds},
  author={Zhao, Lichen and Cai, Daigang and Sheng, Lu and Xu, Dong},
  booktitle=ICCV,
  year={2021},
  pages={2928--2937},
}

@inproceedings{zhu2024scanreason,
  title={Scanreason: Empowering 3d visual grounding with reasoning capabilities},
  author={Zhu, Chenming and Wang, Tai and Zhang, Wenwei and Chen, Kai and Liu, Xihui},
  booktitle=ECCV,
  pages={151--168},
  year={2024},
  organization={Springer}
}

@inproceedings{Yang2023LLMGrounder,
  title={LLM-Grounder: Open-Vocabulary 3D Visual Grounding with Large Language Model as an Agent},
  author={Jianing Yang and Xuweiyi Chen and Shengyi Qian and Nikhil Madaan and Madhavan Iyengar and David F. Fouhey and Joyce Chai},
  booktitle=ICRA,
  year={2023},
  pages={7694-7701},
}

@inproceedings{yuan2024zsvg3d,
  title={Visual Programming for Zero-shot Open-Vocabulary 3D Visual Grounding}, 
  author={Zhihao Yuan and Jinke Ren and Chun-Mei Feng and Hengshuang Zhao and Shuguang Cui and Zhen Li},
  booktitle=CVPR,
  year={2024},
}

@inproceedings{xu2025vlmgrounder,
  title={Vlm-grounder: A vlm agent for zero-shot 3d visual grounding},
  author={Xu, Runsen and Huang, Zhiwei and Wang, Tai and Chen, Yilun and Pang, Jiangmiao and Lin, Dahua},
  booktitle = CORL,
  year={2025}
}

@inproceedings{li2025seeground,
  title     = {SeeGround: See and Ground for Zero-Shot Open-Vocabulary 3D Visual Grounding},
  author    = {Rong Li and Shijie Li and Lingdong Kong and Xulei Yang and Junwei Liang},
  booktitle = CVPR,
  year      = {2025},
}

@inproceedings{jin2025spazer,
  title={SPAZER: Spatial-Semantic Progressive Reasoning Agent for Zero-shot 3D Visual Grounding},
  author={Jin, Zhao and Tu, Rong-Cheng and Liao, Jingyi and Sun, Wenhao and Luo, Xiao and Liu, Shunyu and Tao, Dacheng},
  booktitle = NIPS,
  year={2025}
}

@inproceedings{lin2025seqvlm,
  title={SeqVLM: Proposal-Guided Multi-View Sequences Reasoning via VLM for Zero-Shot 3D Visual Grounding},
  author={Lin, Jiawen and Bian, Shiran and Zhu, Yihang and Tan, Wenbin and Zhang, Yachao and Xie, Yuan and Qu, Yanyun},
  booktitle=ACMMM,
  pages={3094--3103},
  year={2025}
}

@inproceedings{mi2025languagetospace,
  title={Language-to-Space Programming for Training-Free 3D Visual Grounding},
  author={Mi, Boyu and Wang, Hanqing and Wang, Tai and Chen, Yilun and Pang, Jiangmiao},
  booktitle=EMNLP,
  pages={3844--3864},
  year={2025}
}

@inproceedings{zhang2024agent3d0,
  title={Agent3d-zero: An agent for zero-shot 3d understanding},
  author={Zhang, Sha and Huang, Di and Deng, Jiajun and Tang, Shixiang and Ouyang, Wanli and He, Tong and Zhang, Yanyong},
  booktitle=ECCV,
  pages={186--202},
  year={2024},
  organization={Springer}
}

@inproceedings{schult23mask3d,
  title     = {{Mask3D: Mask Transformer for 3D Semantic Instance Segmentation}},
  author    = {Schult, Jonas and Engelmann, Francis and Hermans, Alexander and Litany, Or and Tang, Siyu and Leibe, Bastian},
  booktitle = ICRA,
  year      = {2023}
}

@inproceedings{carion2026sam3,
  title={Sam 3: Segment anything with concepts},
  author={Carion, Nicolas and Gustafson, Laura and Hu, Yuan-Ting and Debnath, Shoubhik and Hu, Ronghang and Suris, Didac and Ryali, Chaitanya and Alwala, Kalyan Vasudev and Khedr, Haitham and Huang, Andrew and others},
  booktitle = ICLR,
  year={2026}
}

@inproceedings{yao2022react,
  title={React: Synergizing reasoning and acting in language models},
  author={Yao, Shunyu and Zhao, Jeffrey and Yu, Dian and Du, Nan and Shafran, Izhak and Narasimhan, Karthik R and Cao, Yuan},
  booktitle=ICLR,
  year={2022}
}

@inproceedings{dai2017scannet,
    title={ScanNet: Richly-annotated 3D Reconstructions of Indoor Scenes},
    author={Dai, Angela and Chang, Angel X. and Savva, Manolis and Halber, Maciej and Funkhouser, Thomas and Nie{\ss}ner, Matthias},
    booktitle = CVPR,
    year = {2017}
}

@article{Qwen3-VL,
      title={Qwen3-VL Technical Report}, 
      author={Shuai Bai and Yuxuan Cai and Ruizhe Chen and Keqin Chen and Xionghui Chen and Zesen Cheng and Lianghao Deng and Wei Ding and Chang Gao and Chunjiang Ge and Wenbin Ge and Zhifang Guo and Qidong Huang and Jie Huang and Fei Huang and Binyuan Hui and Shutong Jiang and Zhaohai Li and Mingsheng Li and Mei Li and Kaixin Li and Zicheng Lin and Junyang Lin and Xuejing Liu and Jiawei Liu and Chenglong Liu and Yang Liu and Dayiheng Liu and Shixuan Liu and Dunjie Lu and Ruilin Luo and Chenxu Lv and Rui Men and Lingchen Meng and Xuancheng Ren and Xingzhang Ren and Sibo Song and Yuchong Sun and Jun Tang and Jianhong Tu and Jianqiang Wan and Peng Wang and Pengfei Wang and Qiuyue Wang and Yuxuan Wang and Tianbao Xie and Yiheng Xu and Haiyang Xu and Jin Xu and Zhibo Yang and Mingkun Yang and Jianxin Yang and An Yang and Bowen Yu and Fei Zhang and Hang Zhang and Xi Zhang and Bo Zheng and Humen Zhong and Jingren Zhou and Fan Zhou and Jing Zhou and Yuanzhi Zhu and Ke Zhu},
	  journal={arXiv preprint arXiv:2511.21631},
      year={2025}
}

@misc{qwen3.5,
    title  = {{Qwen3.5}: Towards Native Multimodal Agents},
    author = {{Qwen Team}},
    month  = {February},
    year   = {2026},
    url    = {https://qwen.ai/blog?id=qwen3.5}
}

@article{3dllm,
 author = {Hong, Yining and Zhen, Haoyu and Chen, Peihao and Zheng, Shuhong and Du, Yilun and Chen, Zhenfang and Gan, Chuang},
 title = {3D-LLM: Injecting the 3D World into Large Language Models},
 journal = {NeurIPS},
 year = {2023},
}

@inproceedings{d3net2022d,
  title={D 3 net: A unified speaker-listener architecture for 3d dense captioning and visual grounding},
  author={Chen, Dave Zhenyu and Wu, Qirui and Nie{\ss}ner, Matthias and Chang, Angel X},
  booktitle=ECCV,
  pages={487--505},
  year={2022},
  organization={Springer}
}

@inproceedings{jain2022bottom,
  title={Bottom up top down detection transformers for language grounding in images and point clouds},
  author={Jain, Ayush and Gkanatsios, Nikolaos and Mediratta, Ishita and Fragkiadaki, Katerina},
  booktitle=ECCV,
  pages={417--433},
  year={2022},
  organization={Springer}
}

@inproceedings{qian2024multi,
  title={Multi-branch collaborative learning network for 3d visual grounding},
  author={Qian, Zhipeng and Ma, Yiwei and Lin, Zhekai and Ji, Jiayi and Zheng, Xiawu and Sun, Xiaoshuai and Ji, Rongrong},
  booktitle=ECCV,
  pages={381--398},
  year={2024},
  organization={Springer}
}

@inproceedings{wu2023eda,
  title={Eda: Explicit text-decoupling and dense alignment for 3d visual grounding},
  author={Wu, Yanmin and Cheng, Xinhua and Zhang, Renrui and Cheng, Zesen and Zhang, Jian},
  booktitle=CVPR,
  pages={19231--19242},
  year={2023}
}

@inproceedings{xu2024pointllm,
  title={Pointllm: Empowering large language models to understand point clouds},
  author={Xu, Runsen and Wang, Xiaolong and Wang, Tai and Chen, Yilun and Pang, Jiangmiao and Lin, Dahua},
  booktitle=ECCV,
  pages={131--147},
  year={2024},
  organization={Springer}
}

@inproceedings{zhu2025llava3D,
  title={Llava-3d: A simple yet effective pathway to empowering lmms with 3d capabilities},
  author={Zhu, Chenming and Wang, Tai and Zhang, Wenwei and Pang, Jiangmiao and Liu, Xihui},
  booktitle=ICCV,
  pages={4295--4305},
  year={2025}
}

@article{huang2024chatscene,
  title={Chat-scene: Bridging 3d scene and large language models with object identifiers},
  author={Huang, Haifeng and Chen, Yilun and Wang, Zehan and Huang, Rongjie and Xu, Runsen and Wang, Tai and Liu, Luping and Cheng, Xize and Zhao, Yang and Pang, Jiangmiao and others},
  journal=NIPS,
  volume={37},
  pages={113991--114017},
  year={2024}
}

@inproceedings{huang20253drs,
  title={3drs: Mllms need 3d-aware representation supervision for scene understanding},
  author={Huang, Xiaohu and Wu, Jingjing and Xie, Qunyi and Han, Kai},
  booktitle=NIPS,
  year={2025}
}

@inproceedings{zheng2025vgllm,
  title={Learning from videos for 3d world: Enhancing mllms with 3d vision geometry priors},
  author={Zheng, Duo and Huang, Shijia and Li, Yanyang and Wang, Liwei},
  booktitle=NIPS,
  year={2025}
}

@inproceedings{zheng2025video3dllm,
  title={Video-3d llm: Learning position-aware video representation for 3d scene understanding},
  author={Zheng, Duo and Huang, Shijia and Wang, Liwei},
  booktitle=CVPR,
  pages={8995--9006},
  year={2025}
}

@inproceedings{mao2025spatiallm,
  title={Spatiallm: Training large language models for structured indoor modeling},
  author={Mao, Yongsen and Zhong, Junhao and Fang, Chuan and Zheng, Jia and Tang, Rui and Zhu, Hao and Tan, Ping and Zhou, Zihan},
  booktitle=NIPS,
  year={2025}
}

@inproceedings{wang2025streambridge,
  title={Streambridge: Turning your offline video large language model into a proactive streaming assistant},
  author={Wang, Haibo and Feng, Bo and Lai, Zhengfeng and Xu, Mingze and Li, Shiyu and Ge, Weifeng and Dehghan, Afshin and Cao, Meng and Huang, Ping},
  booktitle=NIPS,
  year={2025}
}

@article{li2024llavaonevision,
    title={Llava-onevision: Easy visual task transfer},
    author={Li, Bo and Zhang, Yuanhan and Guo, Dong and Zhang, Renrui and Li, Feng and Zhang, Hao and Zhang, Kaichen and Zhang, Peiyuan and Li, Yanwei and Liu, Ziwei and others},
    journal={arXiv:2408.03326},
    year={2024}
}

@inproceedings{xu2025slowfast1.5,
  title={Slowfast-llava-1.5: A family of token-efficient video large language models for long-form video understanding},
  author={Xu, Mingze and Gao, Mingfei and Li, Shiyu and Lu, Jiasen and Gan, Zhe and Lai, Zhengfeng and Cao, Meng and Kang, Kai and Yang, Yinfei and Dehghan, Afshin},
  booktitle=COLM,
  year={2025}
}

@article{touvron2023llama,
    title={Llama: Open and efficient foundation language models},
    author={Touvron, Hugo and Lavril, Thibaut and Izacard, Gautier and Martinet, Xavier and Lachaux, Marie-Anne and Lacroix, Timoth{\'e}e and Rozi{\`e}re, Baptiste and Goyal, Naman and Hambro, Eric and Azhar, Faisal and others},
    journal={arXiv:2302.13971},
    year={2023}
}

@inproceedings{deitke2025molmo,
  title={Molmo and pixmo: Open weights and open data for state-of-the-art vision-language models},
  author={Deitke, Matt and Clark, Christopher and Lee, Sangho and Tripathi, Rohun and Yang, Yue and Park, Jae Sung and Salehi, Mohammadreza and Muennighoff, Niklas and Lo, Kyle and Soldaini, Luca and others},
  booktitle=CVPR,
  pages={91--104},
  year={2025}
}

@inproceedings{liu2024groundingdino,
  title={Grounding dino: Marrying dino with grounded pre-training for open-set object detection},
  author={Liu, Shilong and Zeng, Zhaoyang and Ren, Tianhe and Li, Feng and Zhang, Hao and Yang, Jie and Jiang, Qing and Li, Chunyuan and Yang, Jianwei and Su, Hang and others},
  booktitle=ECCV,
  pages={38--55},
  year={2024},
  organization={Springer}
}

@inproceedings{ester1996dbscan,
  title={A density-based algorithm for discovering clusters in large spatial databases with noise},
  author={Ester, Martin and Kriegel, Hans-Peter and Sander, J{\"o}rg and Xu, Xiaowei and others},
  booktitle={kdd},
  volume={96},
  number={34},
  pages={226--231},
  year={1996}
}

@article{rusu2010sor,
  title={Semantic 3D object maps for everyday manipulation in human living environments},
  author={Rusu, Radu Bogdan},
  journal={KI-K{\"u}nstliche Intelligenz},
  volume={24},
  number={4},
  pages={345--348},
  year={2010},
  publisher={Springer}
}

@inproceedings{wang2024g3lq,
  title={G\^{} 3-lq: Marrying hyperbolic alignment with explicit semantic-geometric modeling for 3d visual grounding},
  author={Wang, Yuan and Li, Yali and Wang, Shengjin},
  booktitle=CVPR,
  pages={13917--13926},
  year={2024}
}

@inproceedings{deng20253dllava,
  title={3d-llava: Towards generalist 3d lmms with omni superpoint transformer},
  author={Deng, Jiajun and He, Tianyu and Jiang, Li and Wang, Tianyu and Dayoub, Feras and Reid, Ian},
  booktitle=CVPR,
  pages={3772--3782},
  year={2025}
}

@inproceedings{qi2026gpt4scene,
  title={Gpt4scene: Understand 3d scenes from videos with vision-language models},
  author={Qi, Zhangyang and Zhang, Zhixiong and Fang, Ye and Wang, Jiaqi and Zhao, Hengshuang},
  booktitle=ICLR,
  year={2026}
}

@inproceedings{yuan2021instancerefer,
  title={Instancerefer: Cooperative holistic understanding for visual grounding on point clouds through instance multi-level contextual referring},
  author={Yuan, Zhihao and Yan, Xu and Liao, Yinghong and Zhang, Ruimao and Wang, Sheng and Li, Zhen and Cui, Shuguang},
  booktitle=ICCV,
  pages={1791--1800},
  year={2021}
}

@inproceedings{zhu20233dvista,
  title={3d-vista: Pre-trained transformer for 3d vision and text alignment},
  author={Zhu, Ziyu and Ma, Xiaojian and Chen, Yixin and Deng, Zhidong and Huang, Siyuan and Li, Qing},
  booktitle=ICCV,
  pages={2911--2921},
  year={2023}
}

@inproceedings{jia2024sceneverse,
  title={Sceneverse: Scaling 3d vision-language learning for grounded scene understanding},
  author={Jia, Baoxiong and Chen, Yixin and Yu, Huangyue and Wang, Yan and Niu, Xuesong and Liu, Tengyu and Li, Qing and Huang, Siyuan},
  booktitle=ECCV,
  pages={289--310},
  year={2024},
  organization={Springer}
}

@article{zhang2025spar,
  title={From flatland to space: Teaching vision-language models to perceive and reason in 3d},
  author={Zhang, Jiahui and Chen, Yurui and Zhou, Yanpeng and Xu, Yueming and Huang, Ze and Mei, Jilin and Chen, Junhui and Yuan, Yu-Jie and Cai, Xinyue and Huang, Guowei and others},
  journal={arXiv preprint arXiv:2503.22976},
  year={2025}
}

\clearpage

\appendix

\begin{figure*}[!t]
    \centering
    \includegraphics[width=1\linewidth]{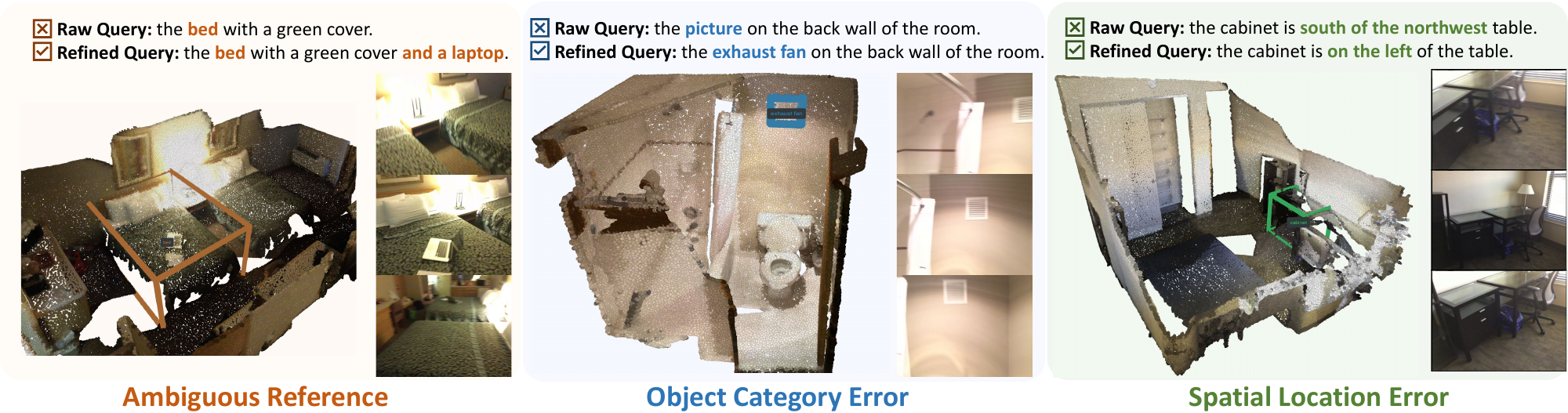}
    \caption{Examples of annotation noise in benchmarks.}
    \label{fig:refine}
    \vspace{-5pt}
\end{figure*}

\section{Appendix}

\subsection{Benchmark Refinement Details}
\label{sec:appendix_benchmark}

While established benchmarks such as ScanRefer \cite{chen2020scanrefer} and Nr3D \cite{achlioptas2020referit3d} have been instrumental in advancing 3D Visual Grounding (3D-VG), we observed that annotation noise within their evaluation subsets inherently penalizes models with precise spatial reasoning capabilities. To ensure a rigorous assessment of our framework and provide a cleaner testbed for future research, we conducted a manual review and refinement process. 

Specifically, we manually audited the 250 sampled queries from the evaluation subsets of each benchmark (500 samples in total). To achieve this, we utilized Open3D to visualize the ground-truth 3D bounding boxes alongside their corresponding point clouds for each sample. This allowed us to manually verify whether the bounding box accurately corresponds to the query, and to determine if the linguistic description is precise or suffers from ambiguity and errors. During this process, we identified and corrected errors that we categorized into three primary types as in Figure \ref{fig:refine}:

\begin{itemize}[leftmargin=*]
    \item \textbf{Ambiguous References:} This is the most prevalent error type. It occurs when a query lacks distinctive features or contextual anchors to uniquely identify a single target, yielding multiple valid candidates within the scene. We resolved these ambiguities by appending exclusive contextual anchors (e.g., modifying ``the bed with a green cover'' to ``the bed with a green cover and a laptop'').
    
    \item \textbf{Spatial Location Errors:} These errors involve queries that describe spatial relationships contradicting the actual 3D physical layout. This often includes erroneous prepositions (e.g., using ``top left'' instead of ``bottom'') or invalid global directions (e.g., referencing ``south'' without a compass reference). We corrected these by substituting contradictory coordinates with reliable, relative spatial anchors.
    
    \item \textbf{Object Category Errors:} These are instances where the class name provided in the query explicitly contradicts the visual reality of the object (e.g., mislabeling an ``exhaust fan'' as a ``picture''). We refined these category names to align with the true visual semantics, enabling accurate semantic anchoring.
\end{itemize}

\subsection{Expert Skill: 3D Visual Grounding}
\label{sec:appendix_skill}
The expert skill serves as the blueprint for the agent, defining the standard operating procedure for the 3D visual grounding and reconstruction pipeline. It is structured as a Markdown document designed to be directly read and parsed by the agent.

\begin{tcolorbox}[breakable, colback=blue!5!white,colframe=blue!75!black,title=Skill: 3D Visual Grounding Pipeline, left=2mm, right=2mm, top=2mm, bottom=2mm, fontupper=\small]

\textbf{Name:} \texttt{3d\_visual\_grounding} \\
\textbf{Description:} Expert Skill for 3D Visual Grounding \& Reconstruction Pipeline. Use this when the user gives you a \texttt{scene\_id} and a \texttt{query}.

\vspace{1mm}
\hrule
\vspace{2mm}

\textbf{When to use this skill:}\\
Use this skill when the user requests to find and reconstruct a specific object in a scene (video with $\sim$300 frames) and calculate its 3D bounding box coordinates.

\vspace{1mm}
\hrule
\vspace{2mm}

\textbf{Instructions:}\\
You MUST strictly follow this sequential execution path:
\begin{enumerate}[leftmargin=*, topsep=0pt, itemsep=2pt]
    \item \textbf{Query Analysis}: Parse the user's natural language query to extract the \texttt{target\_class}, visual \texttt{attributes}, spatial \texttt{conditions}, and global \texttt{scene\_features}.
    \item \textbf{Initialize Image Registry}: Call the read tool to index all image file names from the scene video into a standard JSON list.
    \item \textbf{Coarse Filtering (Object Masks)}: Apply mask-based filtering to retain only images containing the \texttt{target\_class}. \textit{Critical Check}: If 0 images remain, do not proceed. Analyze if the threshold is too high and retry with a lower value.
    \item \textbf{Fine Filtering (Scene Semantics)}: Apply VLM-based filtering to verify if images match the \texttt{scene\_feature}. \textit{Fallback Strategy}: If this filter removes ALL images, discard this step and revert to the result of \textbf{Coarse Filtering (Object Masks)}. Use the images filtered by \textbf{Coarse Filtering (Object Masks)} as the images.
    \item \textbf{Scoring \& Ranking}: Score the remaining images based on their alignment with the query's \texttt{attributes} and \texttt{conditions}. Sort in descending order.
    \item \textbf{Optimal View \& Target Selection}: Traverses the candidate images to find the most distinct observation of the scene as the ``Reference View'', and uses VLM to identify the specific Object ID within this view that best matches the query description.
    \item \textbf{Reference Target Isolation}: Generate a visualization isolating the specific Object ID.
    \item \textbf{Temporal Expansion (Video Tracking)}: Expand the target object search temporally from the Reference View. It tracks the object frame-by-frame (forward and backward) using the VLM to verify identity and SAM to generate masks. This ensures temporal consistency and generates a continuous video clip with the target object. This leverages temporal consistency to robustly collect a continuous video clip containing the target as candidate frames.
    \item \textbf{Candidate Segmentation}: For every verified candidate frame, generate and save the precise segmentation mask for the target object.
    \item \textbf{Initial 3D Reconstruction}: Aggregate all verified images and their masks from the temporal expansion steps to generate an initial, preliminary 3D point cloud.
    \item \textbf{Geometric Multi-View Expansion (Centroid Complete)}: Extract the target's absolute 3D centroid and mathematically project it across other scene frames, utilizing camera extrinsics and Z-Buffer checks.
    \item \textbf{Final Dense 3D Reconstruction}: Reconstruct the updated, comprehensive image and mask paths into a final, dense, and structurally complete PLY point cloud.
    \item \textbf{3D Bounding Box Calculation}: Calculate the axis-aligned 3D bounding box.
\end{enumerate}
\vspace{1mm}
\hrule
\vspace{2mm}
\textbf{Strategy Tips:}\\
\textbf{Dynamic Adjustment}: If a specific step returns 0 valid images, DO NOT give up immediately. Consider relaxing constraints or skipping it.\\
\textbf{Threshold Consistency}: If you lower the \texttt{threshold} in an earlier step, you MUST maintain this lowered threshold for ALL subsequent steps.
\end{tcolorbox}

\subsection{Prompts}
\label{sec:appendix_prompts}

We provide the prompts utilized across different modules of our framework. These prompts dictate the reasoning, filtering, and tracking behaviors of the VLM agent.

\begin{tcolorbox}[breakable, colback=gray!5!white,colframe=gray!75!black,title=Agent System Prompt, left=2mm, right=2mm, top=2mm, bottom=2mm, fontupper=\small]
You are an intelligent assistant capable of calling external tools.

\textbf{\#\#\# YOUR EXPERT SKILLS}\\
\{skill\_description\}

\textbf{\#\#\# AVAILABLE TOOLS}\\
The available tools and their STRICT parameter definitions are as follows:\\
\{tool\_descriptions\}

\textbf{\#\#\# RESPONSE FORMAT}\\
Please respond in a step-by-step loop. For each step, output a Thought followed by an Action, strictly according to the following format:

\textbf{Thought:} 
\begin{itemize}[leftmargin=*, topsep=0pt, itemsep=0pt]
    \item First, recall "YOUR EXPERT SKILLS" to decide which skill to use and the current pipeline stage.
    \item Analyze the previous Observation.
    \item Plan the next step accordingly.
    \item \textbf{CRITICAL}: Check if the previous observation indicates failure. If so, consider retrying with looser constraints or ABORTING.
\end{itemize}

\textbf{Action:} The action you decide to take. It MUST be formatted EXACTLY as one of the following:
\begin{enumerate}[leftmargin=*, topsep=0pt, itemsep=0pt]
    \item \textbf{Tool Call}: \texttt{<tool\_name>(<tool\_input\_json>)}
    \begin{itemize}[leftmargin=*, topsep=0pt, itemsep=0pt]
        \item \texttt{<tool\_name>} must be one of: [\{tool\_names\}]
        \item \texttt{<tool\_input\_json>} must be a valid JSON string containing arguments matching the tool's schema.
        \item Example: \texttt{filter\_by\_mask(\{"threshold": 0.5\})}
    \end{itemize}
    \item \textbf{Finish Task}: \texttt{Finish[<your final answer here>]}
    \begin{itemize}[leftmargin=*, topsep=0pt, itemsep=0pt]
        \item Use this when you have obtained the final answer or completed the task successfully.
    \end{itemize}
    \item \textbf{Abort Task}: \texttt{Abort[<failure\_reason>]}
    \begin{itemize}[leftmargin=*, topsep=0pt, itemsep=0pt]
        \item Use this when the task CANNOT be completed.
    \end{itemize}
\end{enumerate}

\end{tcolorbox}

\begin{tcolorbox}[breakable, colback=gray!5!white,colframe=gray!75!black,title=Query Parsing Prompt, left=2mm, right=2mm, top=2mm, bottom=2mm, fontupper=\small]
You are working on a 3D visual grounding task. You will receive a natural language query that specifies a particular object by describing its attributes and grounding conditions in a scene.

\textbf{Definitions:}
\begin{itemize}[leftmargin=*, topsep=0pt, itemsep=0pt]
    \item \textbf{Target object phrase}: The core noun phrase identifying the object. Unlike a simple class name, this \textbf{must include 1-2 key adjectives} from the query if present (e.g., use "rectangular dark cabinet" instead of just "cabinet"). Include inherent adjectives (color, shape, material). Do NOT include spatial or relative adjectives.
    \item \textbf{Attributes}: Inherent properties of the target object itself, such as category description, color, material, shape, appearance, function, or state.
    \item \textbf{Grounding conditions}: Relational, spatial, or contextual constraints that help uniquely locate the target object relative to other objects, regions, or layouts in the scene.
    \item \textbf{Scene feature}: A single sentence describing the scene's composition based \textbf{STRICTLY} on the objects, regions, and layouts explicitly mentioned in the query. Do \textbf{NOT} infer or hallucinate context.
\end{itemize}

Your task:
1. Parse the query.
2. Identify and return: the target object's phrase, a list of the object's attributes, a list of grounding conditions, and a single string describing the scene feature.

Your response must be formatted strictly as JSON wrapped inside triple backticks:
\begin{verbatim}
{
  "target_class": "...",
  "attributes": [...],
  "conditions": [...],
  "scene_feature": "..."
}
\end{verbatim}
\end{tcolorbox}

\begin{tcolorbox}[breakable, colback=gray!5!white,colframe=gray!75!black,title=Scene Filter Prompt, left=2mm, right=2mm, top=2mm, bottom=2mm, fontupper=\small]
You are a strict visual verification assistant. Your task is to analyze an image and determine if it matches the scene description provided by the user.

You must rigorously verify the image against the user's text across these specific dimensions:
\begin{enumerate}[leftmargin=*, topsep=0pt, itemsep=0pt]
    \item \textbf{Objects}: Confirm that EVERY physical object mentioned in the description is clearly visible. \textbf{IMPORTANT EXCEPTION}: Ignore broad functional area labels or room types (e.g., "living room", "kitchen"). Focus ONLY on specific physical items.
    \item \textbf{Quantity}: Strictly count the objects. If the text specifies a number, the image must show equal to or more than that count.
    \item \textbf{Position \& Space}: Verify the spatial layout, but \textbf{RELAX directional strictness}:
    \begin{itemize}[leftmargin=*, topsep=0pt, itemsep=0pt]
        \item \textbf{Ignore "Left/Right"}: Due to variable camera angles, strictly ignore absolute instructions like "to the left of".
        \item \textbf{Adjacency Rule}: Treat any mention of "left" or "right" simply as \textbf{"next to"} or \textbf{"nearby"}. 
        \item \textbf{Other Spatial Relations}: Maintain strict verification for non-directional spatial terms (e.g., "on top of", "under").
    \end{itemize}
    \item \textbf{Attributes}: Check that key visual attributes (e.g., color, material) match the description.
\end{enumerate}

\textbf{CRITICAL INSTRUCTION}: This is a strict verification task, EXCEPT for the left/right relaxation mentioned above. Only answer "yes" if all constraints are satisfied. Do not provide explanations, just directly answer "yes" or "no".
\end{tcolorbox}

\begin{tcolorbox}[breakable, colback=gray!5!white,colframe=gray!75!black,title=VLM Score Prompt, left=2mm, right=2mm, top=2mm, bottom=2mm, fontupper=\small]
You are a visual filtering assistant. Your goal is to determine if the \textbf{Target Object} described in the user's query is visible in the provided image.

\textbf{\#\#\# Evaluation Steps}
\begin{enumerate}[leftmargin=*, topsep=0pt, itemsep=0pt]
    \item \textbf{Target Detection (The Strict Gatekeeper)}: Focus ONLY on the \texttt{target\_class}. Is this specific object visible in the image? If NO: The score MUST be exactly 0.0.
    \item \textbf{Context Verification}: Check the \texttt{conditions} regarding other objects (Reference Objects). Are the reference objects mentioned visible? If MISSING: The score must be low (0.1 - 1.9).
    \item \textbf{Attribute Matching}: Evaluate visual details (color, shape, state).
\end{enumerate}

\textbf{\#\#\# Confidence Score Scale (Float)}
\begin{itemize}[leftmargin=*, topsep=0pt, itemsep=0pt]
    \item \textbf{0.0}: Target Absent.
    \item \textbf{0.1 - 1.9}: Present, but Context Missing / Strong Mismatch.
    \item \textbf{2.0 - 2.9}: Present, Low Match.
    \item \textbf{3.0 - 3.9}: Present, Partial Match / Ambiguous.
    \item \textbf{4.0 - 4.9}: Present, High Match.
    \item \textbf{5.0}: Present, Perfect Match.
\end{itemize}

Output Format: You must return a strict JSON object containing \texttt{is\_present} (Boolean) and \texttt{score} (Float). Start directly with \texttt{\{} and end with \texttt{\}}.
\end{tcolorbox}

\begin{tcolorbox}[breakable, colback=gray!5!white,colframe=gray!75!black,title=Frame Expansion Prompt, left=2mm, right=2mm, top=2mm, bottom=2mm, fontupper=\small]
You are an expert in Video Object Tracking and Visual Re-Identification.

\textbf{\#\#\# TASK}
Your task is to determine if a specific \textbf{Target Object} (highlighted by bounding boxes in the Reference Video) is present in the provided \textbf{Candidate Image}.

\textbf{\#\#\# CONTEXT}
The Candidate Image is an \textbf{adjacent frame} (temporally continuous) to the Reference Video clip. The object, if present, will have a similar appearance, scale, and location, subject to minor motion or camera movement.

\textbf{\#\#\# JUDGMENT CRITERIA}
\begin{enumerate}[leftmargin=*, topsep=0pt, itemsep=0pt]
    \item \textbf{Identity Consistency}: The object must be the \textbf{exact same instance}, not just a similar category.
    \item \textbf{Robustness}: Allow for slight changes in viewpoint, lighting, scale, or partial occlusion due to motion.
    \item \textbf{Strict Rejection}: If the object is fully occluded or has moved out of the frame, answer NO.
\end{enumerate}

Output strictly \textbf{one word}. If the target is present: "YES". If the target is absent: "NO".
\end{tcolorbox}

\begin{tcolorbox}[breakable, colback=gray!5!white,colframe=gray!75!black,title=Segmentation Marker Prompt, left=2mm, right=2mm, top=2mm, bottom=2mm, fontupper=\small]
You are an expert who identifies objects in annotated scenes. You will receive an Annotated Image with \textbf{bounding boxes} and \textbf{numeric IDs} (0, 1, 2, ...) highlighting the objects.

\textbf{\#\#\# CRITICAL VISUAL GUIDELINES}
\begin{enumerate}[leftmargin=*, topsep=0pt, itemsep=0pt]
    \item \textbf{Focus Inside the Box}: The bounding box delimits the area of interest. To verify \texttt{attributes}, you must analyze the visual content \textbf{strictly within} the boundaries of the box for that ID.
    \item \textbf{Spatial Reasoning (High Priority)}: Pay strict attention to spatial descriptors in the query (e.g., "leftmost," "behind," "next to").
    \item \textbf{Object Matching}: Ensure the object inside the bounding box matches the \texttt{target\_class}.
\end{enumerate}

\textbf{\#\#\# Your Task}
\begin{itemize}[leftmargin=*, topsep=0pt, itemsep=0pt]
    \item \textbf{Single Candidate Override}: If there is ONLY ONE object annotated (only ID 0 exists), strictly output \textbf{ID: 0} immediately.
    \item \textbf{Strict Verification (Multi-object)}: Scan the annotated image to find the object ID that matches the class, spatial conditions, and attributes.
    \item \textbf{Failure Handling}: If NONE satisfy all constraints, output \textbf{ID: -1}.
\end{itemize}

Output ONLY the ID in the format: \texttt{ID: <number>}
\end{tcolorbox}

\subsection{Tool Library}
\label{sec:appendix_tools}
The \textbf{\textit{TAB}} agent is equipped with a comprehensive library of specialized tools. These tools encapsulate foundation vision models (e.g., SAM), Vision-Language Models, and multi-view geometric projection functions. The agent dynamically invokes these tools via strict JSON parameter schemas to execute its planned actions. Table \ref{tab:agent_tools} details the complete tool registry.

\begin{table*}[h]
\centering
\renewcommand{\arraystretch}{1.3}
\begin{tabularx}{\textwidth}{@{} l X @{}}
\toprule
\textbf{Tool Name} & \textbf{Description} \\
\midrule

\texttt{query\_parse()} & Parses the natural language query into a structured JSON format containing \texttt{target\_class}, \texttt{attributes}, \texttt{conditions}, and \texttt{scene\_feature}. \\

\texttt{read\_image\_files()} & Scans the local directory for a specific scene and indexes all image file paths into a structured list. \\

\texttt{object\_filter()} & Filters candidate images using GroundingDINO to retain only frames containing the requested \texttt{target\_class}. \\

\texttt{vlm\_filter()} & Utilizes a Vision-Language Model to verify if an image strictly satisfies the global scene constraints. \\

\texttt{vlm\_score()} & Scores and ranks candidate images based on how well their visual contents match the query's attributes and spatial conditions. \\

\texttt{argmax\_image\_and\_seg\_id()} & Select the best candidate image and identify the specific target object ID within it. It iterates through the images (from highest score), generates segmentation masks using the SAM, and uses the VLM to pinpoint the specific object ID matching the query. \\

\texttt{segment\_target\_in\_reference()} & Isolate the specific target object in the reference image. It draws a clean bounding box around the identified target ID to create a 'Reference View'. \\

\texttt{vlm\_frame\_expansion()} & Expand the target object search temporally from a reference frame. It tracks the object frame-by-frame (forward and backward) using the VLM to verify identity and the SAM to generate masks. \\

\texttt{segment\_all\_target\_object()} & Perform segmentation on all candidate images for reconstruction. It iterates through the list of validated images, generates segmentation masks, and uses the VLM to identify and save the specific mask corresponding to the target object in each view. \\

\texttt{reconstruct\_point\_cloud()} & Generates the 3D point cloud by lifting the segmented target images back into 3D space using camera parameters. \\

\texttt{centroid\_complete()} & Extracts the target's 3D centroid and mathematically projects it across unobserved frames with depth-based occlusion checks, maximizing view coverage. \\

\texttt{calculate\_bbox()} & Calculates and outputs the final axis-aligned 3D bounding box from the reconstructed target point cloud. \\

\bottomrule
\end{tabularx}
\caption{The complete registry of specialized tools available to the agent.}
\label{tab:agent_tools}
\end{table*}

\subsection{Agent Execution Trace Example}
\label{sec:appendix_trace}

To concretely demonstrate how the framework operates in practice, we provide a complete execution trace for a complex query. The agent successfully follows the \textit{3D Visual Grounding} skill pipeline, from initial query parsing to the final 3D bounding box calculation. 

\textbf{Scene ID:} \texttt{scene0435\_00} \\
\textbf{Query:} \textit{``the pillow on the left bed. it is the top pillow on the side of the bed that is closer to the table between the beds.''}

\vspace{0.5em}

\begin{tcolorbox}[breakable, colback=white,colframe=gray!75!black,title=Execution Trace, breakable, left=2mm, right=2mm, top=2mm, bottom=2mm, fontupper=\small]

\textbf{\textcolor{blue}{[Step 1] Query Analysis}} \\
\textbf{Thought:} 
\begin{itemize}[leftmargin=*, topsep=0pt, itemsep=0pt, parsep=0pt]
    \item I need to begin the 3D Visual Grounding \& Reconstruction Pipeline as per the instructions.
    \item The first step is to parse the user's query to extract structured information: \texttt{target\_class}, \texttt{attributes}, \texttt{conditions}, and \texttt{scene\_feature}.
    \item This will guide all subsequent steps in the pipeline.
\end{itemize}
\textbf{Action:} \texttt{query\_parse(\{"query": "..."\})} \\
\textbf{Observation:} 
\begin{verbatim}
{
  "target_class": "pillow",
  "attributes": ["top pillow"],
  "conditions": ["it's on the left bed", 
                "it's the top pillow", 
  "it's on the side... closer to the table..."],
  "scene_feature": "The scene contains two beds 
                with a table between them..."
}
\end{verbatim}

\vspace{0.5em} \hrule \vspace{0.5em}

\textbf{\textcolor{blue}{[Step 2] Initialize Image Registry}} \\
\textbf{Thought:} 
\begin{itemize}[leftmargin=*, topsep=0pt, itemsep=0pt, parsep=0pt]
    \item The query has been successfully parsed into structured JSON, extracting the target class (pillow), conditions, and scene feature.
    \item The next step is to initialize the image registry by reading all image file paths from the scene directory.
    \item This will provide the full set of frames to work with for subsequent filtering and reconstruction steps.
\end{itemize}
\textbf{Action:} \texttt{read\_image\_files(\{"scene\_id": "scene0435\_00"\})} \\
\textbf{Observation:} \texttt{image\_files.json}

\vspace{0.5em} \hrule \vspace{0.5em}

\textbf{\textcolor{blue}{[Step 3] Coarse Filtering}} \\
\textbf{Thought:} 
\begin{itemize}[leftmargin=*, topsep=0pt, itemsep=0pt, parsep=0pt]
    \item The image registry has been initialized successfully with all image paths stored in 'image\_files.json'.
    \item The next step is to perform Coarse Filtering using mask-based detection to retain only images containing the target class ('pillow').
    \item I will use the default threshold of 0.5 for mask generation. If no images are found, I will retry with a lower threshold as per the failure strategy.
\end{itemize}
\textbf{Action:} \texttt{object\_filter(\{"threshold": 0.5, ...\})} \\
\textbf{Observation:} \texttt{object\_filtered\_image\_files.json, 90} (Images retained)

\vspace{0.5em} \hrule \vspace{0.5em}

\textbf{\textcolor{blue}{[Step 4] Fine Filtering}} \\
\textbf{Thought:} 
\begin{itemize}[leftmargin=*, topsep=0pt, itemsep=0pt, parsep=0pt]
    \item Coarse Filtering successfully returned 90 images containing the target class 'pillow'.
    \item The next step is Fine Filtering using VLM to verify if these images match the scene\_feature: "The scene contains two beds with a table between them, and the left bed has a pillow on the side closer to the table."
    \item This step will help narrow down images that match the broader scene context.
\end{itemize}
\textbf{Action:} \texttt{vlm\_filter(\{...\})} \\
\textbf{Observation:} \texttt{vlm\_filtered\_image\_files.json, 24} (Images retained)

\vspace{0.5em} \hrule \vspace{0.5em}

\textbf{\textcolor{blue}{[Step 5] Scoring \& Ranking}} \\
\textbf{Thought:} 
\begin{itemize}[leftmargin=*, topsep=0pt, itemsep=0pt, parsep=0pt]
    \item VLM-based Fine Filtering returned 24 images that match the scene feature.
    \item The next step is to score and rank these images based on their alignment with the full query (including attributes and conditions).
    \item This will help prioritize the most relevant views for subsequent steps.
\end{itemize}
\textbf{Action:} \texttt{vlm\_score(\{...\})} \\
\textbf{Observation:} \texttt{('vlm\_scores.json', 'vlm\_ranked\_image\_files.json')}

\vspace{0.5em} \hrule \vspace{0.5em}

\textbf{\textcolor{blue}{[Step 6] Optimal View \& Target Selection}} \\
\textbf{Thought:} 
\begin{itemize}[leftmargin=*, topsep=0pt, itemsep=0pt, parsep=0pt]
    \item The images have been successfully scored and ranked based on relevance to the query.
    \item The next step is to select the best candidate image and identify the specific target object ID within it.
    \item I will use the highest-scoring image and apply the same threshold (0.5) used before for consistency.
\end{itemize}
\textbf{Action:} \texttt{argmax\_image\_and\_seg\_id(\{...\})} \\
\textbf{Observation:} \texttt{('.../00085.jpg', 5, 'reference\_image\_mask\_results.pth')}

\vspace{0.5em} \hrule \vspace{0.5em}

\textbf{\textcolor{blue}{[Step 7] Reference Target Isolation}} \\
\textbf{Thought:} 
\begin{itemize}[leftmargin=*, topsep=0pt, itemsep=0pt, parsep=0pt]
    \item The best candidate image has been identified: '.../00085.jpg'.
    \item The specific target object ID is 5.
    \item The next step is to isolate the target object in this reference image by generating a clean visualization with a bounding box.
\end{itemize}
\textbf{Action:} \texttt{segment\_target\_in\_reference(\{...\})} \\
\textbf{Observation:} 

\texttt{reference\_image\_with\_target\_mask\_box.png}

\vspace{0.5em} \hrule \vspace{0.5em}

\textbf{\textcolor{blue}{[Step 8] Temporal Expansion}} \\
\textbf{Thought:} 
\begin{itemize}[leftmargin=*, topsep=0pt, itemsep=0pt, parsep=0pt]
    \item The reference image with the target pillow isolated and masks has been successfully generated.
    \item The next step is to perform temporal expansion from this reference frame to track the target object across frames.
    \item I will use a maximum of 32 frames for expansion and maintain the threshold of 0.5 for consistency.
\end{itemize}
\textbf{Action:} \texttt{vlm\_frame\_expansion(\{...\})} \\
\textbf{Observation:} \texttt{('expanded\_image\_files.json', 10)}

\vspace{0.5em} \hrule \vspace{0.5em}

\textbf{\textcolor{blue}{[Step 9] Candidate Segmentation}} \\
\textbf{Thought:} 
\begin{itemize}[leftmargin=*, topsep=0pt, itemsep=0pt, parsep=0pt]
    \item Temporal expansion from the reference frame yielded 10 images for 3D reconstruction.
    \item I will proceed to segment the target object in all candidate images to prepare for reconstruction.
    \item This step will generate masks for each image, which are necessary for the reconstruction.
\end{itemize}
\textbf{Action:} \texttt{segment\_all\_target\_object(\{...\})} \\
\textbf{Observation:} \texttt{('final\_images.json', 'final\_masks.pth')} \\

\vspace{0.5em} \hrule \vspace{0.5em}

\textbf{\textcolor{blue}{[Step 10] Initial 3D Reconstruction}} \\
\textbf{Thought:} 
\begin{itemize}[leftmargin=*, topsep=0pt, itemsep=0pt, parsep=0pt]
    \item The target object has been segmented in all candidate images, producing final image and mask files.
    \item The next step is to perform initial 3D reconstruction using these images and masks.
    \item This will create a preliminary point cloud for the target pillow.
\end{itemize}
\textbf{Action:} \texttt{reconstruct\_point\_cloud(\{...\})} \\
\textbf{Observation:} \texttt{pred\_pcd.ply}

\vspace{0.5em} \hrule \vspace{0.5em}

\textbf{\textcolor{blue}{[Step 11] Geometric Multi-View Expansion}} \\
\textbf{Thought:} 
\begin{itemize}[leftmargin=*, topsep=0pt, itemsep=0pt, parsep=0pt]
    \item The initial 3D point cloud has been successfully reconstructed and saved as 'pred\_pcd.ply'.
    \item The next critical step is to perform Geometric Multi-View Expansion using the centroid\_complete tool to maximize view coverage.
    \item This will project the 3D centroid across other scene frames and harvest additional unoccluded views for a more complete reconstruction.
\end{itemize}
\textbf{Action:} \texttt{centroid\_complete(\{...\})} \\
\textbf{Observation:} \texttt{('centroid\_final\_images.json', 'centroid\_final\_masks.pth')}

\vspace{0.5em} \hrule \vspace{0.5em}

\textbf{\textcolor{blue}{[Step 12] Final Reconstruction}} \\
\textbf{Thought:} 
\begin{itemize}[leftmargin=*, topsep=0pt, itemsep=0pt, parsep=0pt]
    \item The centroid\_complete tool has successfully expanded the view coverage, producing a more comprehensive set of images and masks.
    \item The next step is to perform the final dense 3D reconstruction using these updated inputs.
    \item This will generate a denser and more structurally complete point cloud.
\end{itemize}
\textbf{Action:} \texttt{reconstruct\_point\_cloud(\{...\})} \\
\textbf{Observation:} \texttt{pred\_pcd.ply}

\vspace{0.5em} \hrule \vspace{0.5em}

\textbf{\textcolor{blue}{[Step 13] 3D Bounding Box Calculation}} \\
\textbf{Thought:} 
\begin{itemize}[leftmargin=*, topsep=0pt, itemsep=0pt, parsep=0pt]
    \item The final dense 3D point cloud has been successfully reconstructed and saved as 'pred\_pcd.ply'.
    \item The next and final step is to calculate the axis-aligned 3D bounding box of this point cloud.
    \item This will provide the required coordinates for the target pillow.
\end{itemize}
\textbf{Action:} \texttt{calculate\_bbox(\{"pcd\_path": "pred\_pcd.ply"\})} \\
\textbf{Observation:} \texttt{[1.6515, 1.1065, 0.7770, 0.4687, 0.6466, 0.2580]}

\vspace{0.5em} \hrule \vspace{0.5em}

\textbf{\textcolor{blue}{[Final Output]}} \\
\textbf{Thought:} 
\begin{itemize}[leftmargin=*, topsep=0pt, itemsep=0pt, parsep=0pt]
    \item The 3D bounding box has been successfully calculated for the point cloud of the target pillow.
    \item The bounding box coordinates are: [cx cy cz dx dy dz] = [1.6515, 1.1065, 0.7770, 0.4687, 0.6466, 0.2580].
    \item This completes the entire 3D Visual Grounding \& Reconstruction Pipeline.
\end{itemize}
\textbf{Action:} \texttt{Finish[The 3D bounding box for the pillow on the left bed is [1.6515, 1.1065, 0.7770, 0.4687, 0.6466, 0.2580].]}
\end{tcolorbox}

While the complete trace above illustrates a perfectly smooth execution, it is important to note that our framework is designed with strong robustness and dynamic fallback mechanisms to handle challenging or ambiguous observations. If the agent encounters an error or invalid state during an intermediate step, it does not simply crash; instead, it gracefully falls back to a previous stable state. For instance, if the \textit{Geometric Multi-View Expansion} step calculates an invalid \texttt{NaN} centroid due to depth sensor noise or errors, the framework will automatically abort this specific expansion and proceed to perform the 3D reconstruction directly using the valid frames already gathered during the \textit{Temporal Expansion} phase. 

Furthermore, the agent actively self-corrects during semantic and visual filtering through \textbf{Dynamic Adjustment} and \textbf{Threshold Consistency} strategies. If a strict initial constraint (e.g., a default segmentation threshold of 0.5) results in zero valid candidates, the agent does not immediately abort the task. Instead, it dynamically relaxes the parameter to recover the target, and strictly maintains this updated threshold across all subsequent tracking and segmentation modules to prevent logical contradictions. To demonstrate this intelligent self-correction, the following trace snippet showcases the agent recovering from a failed \textit{Coarse Filtering} step and preserving the modified context:

\begin{tcolorbox}[breakable, colback=white,colframe=gray!75!black,title=Fallback Execution Trace, breakable, left=2mm, right=2mm, top=2mm, bottom=2mm, fontupper=\small]

\textbf{\textcolor{blue}{[... Initial Steps Skipped ...]}}

\vspace{0.5em} \hrule \vspace{0.5em}

\textbf{\textcolor{blue}{[Step 3] Coarse Filtering (Object Masks)}} \\
\textbf{Thought:} 
\begin{itemize}[leftmargin=*, topsep=0pt, itemsep=0pt, parsep=0pt]
    \item The image registry has been initialized successfully with all image paths stored in 'image\_files.json'.
    \item The next step is to perform Coarse Filtering using object-based detection to retain only images containing the target class ('small black remote').
    \item I will use the default threshold of 0.5.
\end{itemize}
\textbf{Action:} \begingroup \ttfamily object\_filter(\{"image\_files\_path": "image\_files.json", "parsed\_query": "...", "threshold": 0.5\}) \endgroup \\
\textbf{Observation:} \begingroup \ttfamily object\_filtered\_image\_files.json, 0 \endgroup

\vspace{0.5em} \hrule \vspace{0.5em}

\textbf{\textcolor{blue}{[Step 3] Coarse Filtering (Object Masks) - Retry}} \\
\textbf{Thought:} 
\begin{itemize}[leftmargin=*, topsep=0pt, itemsep=0pt, parsep=0pt]
    \item \textbf{CRITICAL:} The previous filtering step returned 0 valid images. I MUST NOT give up immediately. 
    \item Following the \textbf{Dynamic Adjustment} strategy, the threshold of 0.5 might be too strict for this specific small object. I will retry with a looser threshold of 0.3.
    \item I must also remember to maintain this 0.3 threshold for all subsequent steps according to the \textbf{Threshold Consistency} rule.
\end{itemize}
\textbf{Action:} \begingroup \ttfamily object\_filter(\{"image\_files\_path": "image\_files.json", "parsed\_query": "...", "threshold": 0.3\}) \endgroup \\
\textbf{Observation:} \begingroup \ttfamily object\_filtered\_image\_files.json, 24 \endgroup

\vspace{0.5em} \hrule \vspace{0.5em}

\textbf{\textcolor{blue}{[... Intermediate Steps Skipped ...]}}

\vspace{0.5em} \hrule \vspace{0.5em}

\textbf{\textcolor{blue}{[Step 6] Optimal View \& Target Selection}} \\
\textbf{Thought:} 
\begin{itemize}[leftmargin=*, topsep=0pt, itemsep=0pt, parsep=0pt]
    \item The images have been successfully scored and ranked.
    \item I will select the best candidate image and identify the specific target object ID within it.
    \item \textbf{CRITICAL:} Following the \textbf{Threshold Consistency} rule, I MUST use the lowered threshold of 0.3 that was established during Coarse Filtering, rather than reverting to the default 0.5.
\end{itemize}
\textbf{Action:} \begingroup \ttfamily argmax\_image\_and\_seg\_id(\{"scores\_path": "...", "image\_files\_path": "...", "query": "...", "parsed\_query": "...", "threshold": 0.3\}) \endgroup \\
\textbf{Observation:} \begingroup \ttfamily ('.../00112.jpg', 2, 'reference\_image\_mask\_results.pth') \endgroup

\end{tcolorbox}

\end{document}